\documentclass[]{bytedance_seed}

\usepackage[toc,page,header]{appendix}

\usepackage{minitoc}
\usepackage{enumitem}

\usepackage{pifont}
\usepackage{amssymb}
\usepackage{makecell}
\usepackage{etoc} 
\usepackage{floatflt}
\usepackage{float}

\usepackage{xcolor}
\usepackage{booktabs}
\usepackage{hyperref}

\usepackage{booktabs}
\usepackage{hyperref}
\definecolor{first}{RGB}{200,240,255}    
\newcommand{\first}[1]{\cellcolor{first}{#1}}
\newcommand{\second}[1]{\textbf{#1}}

\usepackage{graphicx}
\usepackage{subcaption}
\usepackage{array}

\usepackage{latexsym}

\usepackage{tabularx}
\usepackage{graphicx} 
\usepackage{booktabs}
\usepackage{multirow} 
\usepackage{colortbl} 
\usepackage{tabularx}
\usepackage{makecell}
\usepackage{booktabs}
\usepackage{listings}

\usepackage{tcolorbox}
\tcbuselibrary{breakable, skins}
\usepackage{tikz}
\usepackage{eurosym}
\definecolor{headercolor}{gray}{0.88}
\newtcolorbox[auto counter, number within=section]{Prompt}[2][]{%
  colback=white, 
  colframe=cyan, 
  width=\textwidth, 
  arc=5mm, 
  boxrule=0.8mm, 
  title=\large #2, 
  breakable, 
  fonttitle=\small, 
  fontupper=\footnotesize, 
  #1 
}

\newtcolorbox[auto counter, number within=section]{QuestionCase}[2][]{%
  colback=white,
  colframe=yellow!50!red,
  width=\textwidth, 
  arc=5mm, 
  boxrule=0.8mm, 
  title=\large #2, 
  breakable, 
  fonttitle=\small, 
  fontupper=\footnotesize, 
  #1 
}
\newtcolorbox[auto counter, number within=section]{TaskCase}[2][]{%
  colback=white,
  colframe=teal!70!black,
  width=\textwidth,
  arc=3mm,
  boxrule=0.6mm,
  title={\textbf{Task \thetcbcounter:} #2},
  breakable,
  fonttitle=\small,
  fontupper=\footnotesize,
  #1
}

\newtcolorbox{EvalMetrics}[1][]{%
  colback=gray!5,
  colframe=gray!50!black,
  width=\textwidth,
  arc=2mm,
  boxrule=0.4mm,
  title={\textbf{Evaluation Metrics}},
  breakable,
  fonttitle=\scriptsize,
  fontupper=\scriptsize,
  #1
}
\usepackage{graphicx}
\usepackage{float}
\usepackage{wrapfig}

\usepackage{booktabs} 
\usepackage{tabularx} 

\usepackage[svgnames]{xcolor} 

\usepackage{hyperref}
\hypersetup{
    colorlinks=true,
    linkcolor=MidnightBlue,
    urlcolor=MidnightBlue,
    citecolor=MidnightBlue,
}
\usepackage{tabularx}
\usepackage{colortbl}
\newtcolorbox{apibox}[1]{
    enhanced,
    attach boxed title to top left={yshift=-2mm, xshift=3mm},
    colback=black!5,          
    colframe=black!60,        
    fonttitle=\bf\texttt,      
    boxrule=0.5pt,            
    title=#1,                 
    sharp corners,
    breakable,                
}

\title{WorldTravel: A Realistic Multimodal Travel-Planning Benchmark with Tightly Coupled Constraints}
\affiliation[1]{ByteDance Seed}
\affiliation[2]{Humanlaya Data}
\contribution{Full author list in Contributions}

\abstract{
Real-world autonomous planning requires coordinating tightly coupled constraints where a single decision dictates the feasibility of all subsequent actions. 
However, existing benchmarks predominantly feature loosely coupled constraints solvable through local greedy decisions and rely on idealized data, failing to capture the complexity of extracting parameters from dynamic web environments. 
We introduce \textbf{WorldTravel}, a benchmark comprising 150 real-world travel scenarios across 5 cities that demand navigating an average of 15+ interdependent temporal and logical constraints. 
To evaluate agents in realistic deployments, we develop \textbf{WorldTravel-Webscape}, a multi-modal environment featuring over 2,000 rendered webpages where agents must perceive constraint parameters directly from visual layouts to inform their planning. 
Our evaluation of 10 frontier models reveals a significant performance collapse: even the state-of-the-art GPT-5.2 achieves only 32.67\% feasibility in text-only settings, which plummets to 19.33\% in multi-modal environments. 
We identify a critical Perception-Action Gap and a Planning Horizon threshold at approximately 10 constraints where model reasoning consistently fails, suggesting that perception and reasoning remain independent bottlenecks. 
These findings underscore the need for next-generation agents that unify high-fidelity visual perception with long-horizon reasoning to handle brittle real-world logistics.
}

\date{\today}
\correspondence{Zhoufutu Wen at \email{liniuniu@bytedance.com}}

\begin{document}
\maketitle


\section{Introduction}\label{sec:intro}

Large Language Models (LLMs) are rapidly transcending their origins as conversational assistants, evolving into autonomous agents that navigate the intricacies of our daily lives~\citep{agent-survey,LD-Agent,ReAct}.
As their functional boundaries expand, we are increasingly delegating tasks that are fundamentally Operations Research (OR) problems in disguise: from decomposing quarterly sales targets into actionable milestones to orchestrating complex, multi-city travel itineraries. 
In these roles, the model is no longer just processing text; it is solving a high-dimensional puzzle of interdependent decisions where every action must align with rigid temporal and logical constraints.
To make this coupling concrete, consider a scenario in one of the world's most logistically demanding cities:


\begin{quote}
\textit{``A 10:00 Louvre entry and a 13:00 gallery closure leave only a narrow lunch window before a hard 15:20 Eurostar cutoff. A small delay or a sold-out reservation can cascade into an infeasible itinerary.''}
\end{quote}

\begin{floatingfigure}[r]{0.5\textwidth}
  \centering
  \includegraphics[width=0.48\textwidth]{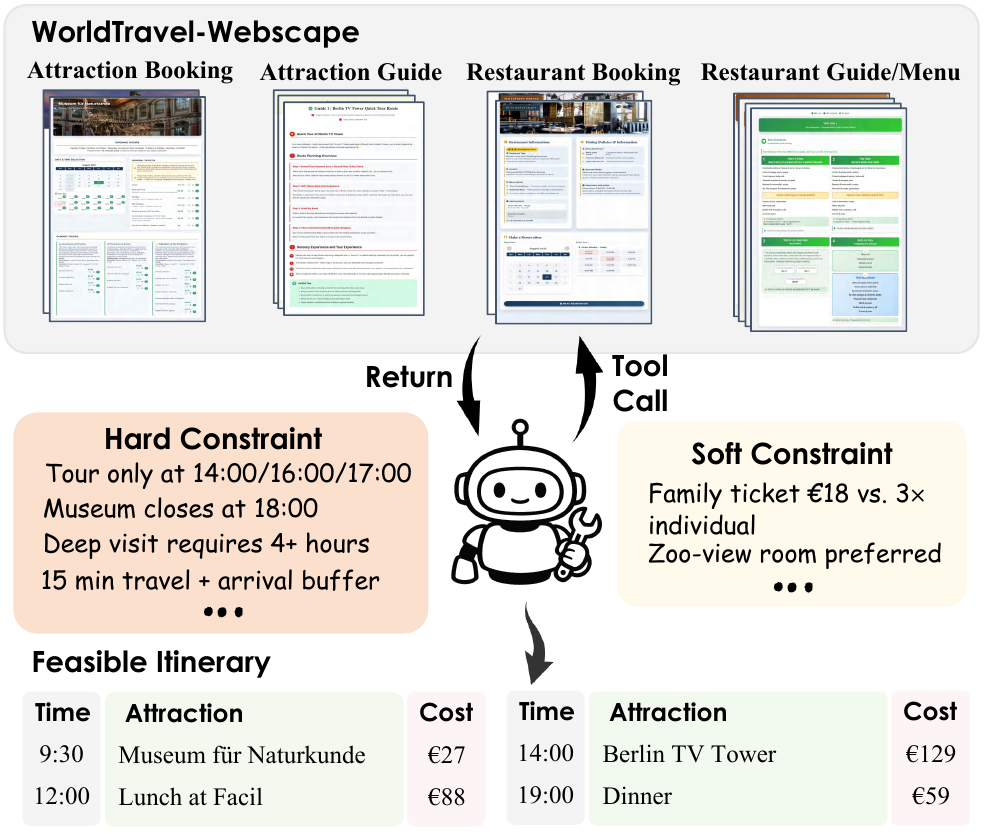}
  \caption{\textbf{Overview of WorldTravel.} Agents interact with \textbf{WorldTravel-Webscape} through tool calls to retrieve constraint parameters from rendered webpages (e.g., attraction booking, restaurant guides). The webpages mirror real booking platforms with realistic UI layouts and interaction patterns (see Appendix~\ref{sec:appendix_web_specs} for specifications). The agent must satisfy hard constraints (timed-entry slots, operating hours, dwell times, travel buffers) and soft constraints (cost optimization, lodging preferences) to produce a feasible itinerary.}
  \label{fig:overall}
\end{floatingfigure}

This brittleness, where a single violation can invalidate the entire plan, is precisely what autonomous agents must handle in real deployments.
Yet despite the high value of such applications, current evaluation frameworks provide only a partial view of these skills.
Existing benchmarks, such as \textit{TravelPlanner}~\citep{tp} and \textit{ChinaTravel}~\citep{chinatravel}, often simplify these challenges into independent, ``greedy-solvable'' subproblems where a local failure does not necessarily compromise global feasibility. 
Furthermore, most existing benchmarks operate in ``clean'' settings where constraint information is already normalized into structured formats (e.g., JSON).
In realistic deployments, the same constraints are rarely explicit or centralized; they must be recovered from rendered webpages and complex UI layouts, including signals like ``Sold Out'' status and tiered pricing.
A natural question, then, is \textit{whether current agents can handle the tightly coupled constraints present in real-world web interfaces}.

To rigorously evaluate end-to-end itinerary planning under realistic execution constraints, we introduce \textbf{WorldTravel}, a benchmark that couples a task suite, a web-interactive evaluation environment, and expert-designed rubrics.
\textbf{WorldTravel} comprises 150 real-world tasks distributed across five major European cities.
Alongside these tasks, we release \textbf{WorldTravel-Webscape}, a controlled interaction space integrating more than 2,000 webpages that mirror real booking platforms. The environment spans 36 attractions, 25 restaurants, and 26 hotels, enabling agents to retrieve constraint parameters via tool calls during execution.
By preserving core sources of planning failure such as tightly coupled temporal constraints and fragmented information while enforcing a deterministic notion of feasibility, \textbf{WorldTravel-Webscape} ensures both answer convergence and reliable evaluation across different agent architectures.
Finally, expert-designed rubrics for each task enable a fine-grained diagnosis of intermediate decisions and specific failure modes.

Our evaluation of 10 frontier models reveals a substantial gap between reasoning and perception. In a text-only setting where parameters are pre-extracted, the best-performing model, GPT-5.2, achieves a feasibility rate of only 32.67\%. When required to extract parameters from rendered webpages in our multi-modal setting, performance drops to 19.33\%. This 13.34 percentage point decline highlights a bottleneck we term the \textit{Perception-Action Gap}, indicating that current models struggle to acquire constraint parameters from visual inputs.

Through ablation studies, we identify a ``Planning Horizon'' effect: model performance degrades consistently as the number of constraints increases, with a clear inflection point at approximately 10 constraints. We further show that visual constraint acquisition and logical reasoning represent two independent bottlenecks. Even when models are provided with pre-extracted parameters in structured form, the best model achieves only 36.67\% feasibility, suggesting that perception and reasoning require separate improvements.

Our contributions are as follows:
\begin{itemize}[noitemsep, topsep=0pt]
    \item We introduce \textbf{WorldTravel}, a benchmark for end-to-end itinerary planning, comprising 150 real-world scenarios across five European cities with tightly coupled temporal and cost constraints.
    
    \item We design \textbf{WorldTravel-Webscape}, a controlled yet realistic multi-modal evaluation environment where agents must extract constraint parameters from rendered webpages; paired with expert-designed rubrics for each task, it enables joint assessment of visual perception and global planning under verifiable feasibility.
    
    \item Through experiment, we identify two fundamental bottlenecks in current agents—the \textbf{Perception-Action Gap} and a \textbf{10-constraint Planning Horizon}—providing concrete diagnostic targets for future research.
\end{itemize}
\section{Related Work}
\label{sec:related_work}

\textbf{LLM-Based Agents.} Large language model-based agents have demonstrated remarkable capabilities in executing multi-step tasks~\citep{agent-survey}, showing potential across domains such as education~\citep{simclass}, healthcare~\citep{huatuogpt}, and long-term personalized dialogue agents~\citep{LD-Agent}. To enhance their planning ability, some works propose decomposing complex tasks into subtasks~\citep{huggingGPT,planandsolve}, while others focus on step-by-step reasoning strategies for dynamic planning~\citep{CoT,GoT,ToT,ReAct}. Additionally, by incorporating long-term memory mechanisms~\citep{MemoryBank,LD-Agent} and external tool use~\citep{Toolformer,ToolLLM}, LLM-based agents can leverage historical knowledge and external resources to significantly extend their problem-solving capacity. However, although these strategies have achieved promising results on certain tasks, their effectiveness in complex scenarios with multiple constraints remains uncertain~\citep{tp}.
\begin{figure*}[!t]
\centering
\includegraphics[width=\linewidth]{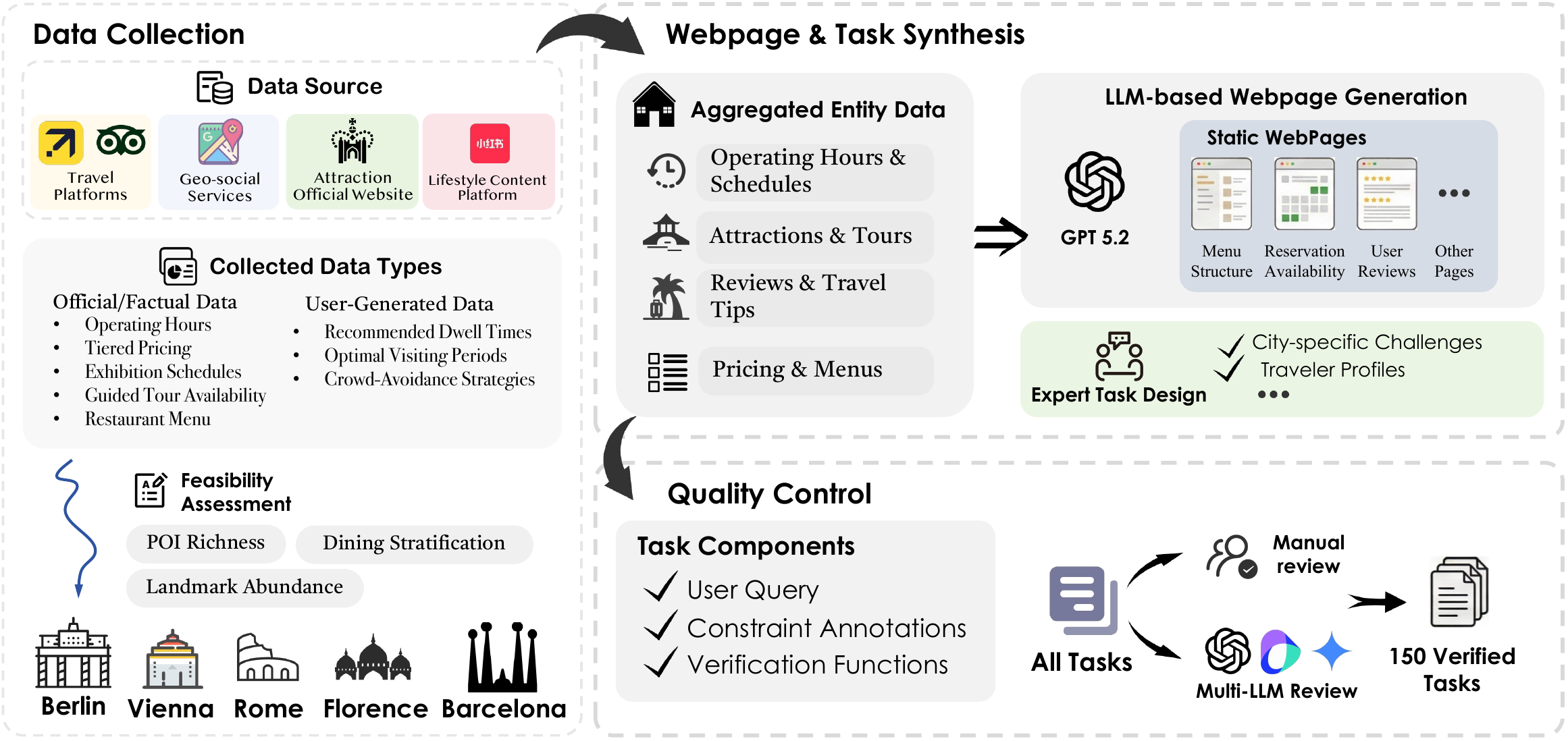}
\caption{\textbf{Data construction pipeline of WorldTravel.} \textbf{(1) Data Collection:} We collect factual data (operating hours, pricing, schedules) and user-generated data (dwell times, visiting tips) from diverse sources across 5 European cities selected based on POI richness. \textbf{(2) Webpage \& Task Synthesis:} Aggregated entity data is used to generate static webpages via GPT-5.2, while expert annotators design tasks with city-specific challenges. \textbf{(3) Quality Control:} Each task includes user queries, constraint annotations, and verification functions; tasks undergo manual review and multi-LLM filtering to yield 150 verified tasks.}
\label{fig:pipeline}
\end{figure*}

\textbf{Travel Planning.} Travel itinerary planning is a complex, multi-constraint task that involves selecting points of interest (POIs), planning spatiotemporal routes, managing budgets, and accommodating personalized preferences~\citep{survey4trevalplan}. Recent benchmarks emphasize realistic scenarios and constraint-driven evaluation for this challenge. TravelPlanner~\citep{tp} is a benchmark that evaluates tool-using language agents on multi-day travel plans under multiple constraints, such as temporal schedules and user preferences. ChinaTravel~\citep{chinatravel} extends this by providing an open-ended benchmark in authentic Chinese travel scenarios, leveraging a domain-specific language (DSL) to compositionally check logical constraints and preference satisfaction. Travel-Sim~\citep{Treval-Sim} introduces an agent-based simulator to assess whether a generated itinerary remains feasible when “executed” in a step-by-step travel simulation, thus evaluating long-horizon plan robustness. In parallel, new system designs focus on reliability and personalization in itinerary planning. TravelAgent~\citep{TravelAgent} is an LLM-powered assistant that integrates information retrieval via tool use, POI recommendation, step-by-step planning, and a memory module to produce rational and customized multi-day itineraries. ItiNera~\citep{ItiNera} combines large language models with cluster-based spatial optimization to generate itineraries that are personalized yet spatially coherent (efficient routing) in urban settings. A complementary neuro-symbolic approach~\citep{hao} translates natural-language travel requirements into formal satisfiability (SMT) constraints and employs a solver to generate plans that are provably correct with respect to all given constraints. Each of these frameworks addresses different aspects of the travel planning problem, pushing the capabilities of AI planners under intertwined real-world constraints.
\section{The WorldTravel Benchmark}
\label{sec:method}

\begin{table*}[t]
\centering
\caption{\textbf{Constraint taxonomy in WorldTravel.} Hard constraints ($\mathcal{C}_H$) govern temporal feasibility and must be strictly satisfied. Any violation renders the itinerary infeasible. Soft constraints ($\mathcal{C}_S$) capture decision correctness (cost accounting, lodging selection) and are evaluated only within feasible solutions.}
\label{tab:constraints}
\resizebox{!}{\height}{%
\small
\begin{tabular}{@{}p{3.3cm}p{10cm}p{2.5cm}@{}}
\toprule
\textbf{Constraint Type} & \textbf{Description} & \textbf{Definition} \\
\midrule
\rowcolor{headercolor}
\multicolumn{3}{@{}l}{\textbf{Temporal Feasibility Constraints} (Hard Constraints $\mathcal{C}_H$)} \\
Timed-Entry Slot & For attractions and restaurants requiring reservations, the activity must start at an available discrete time slot shown on the booking page (e.g., 14:00 guided tour, 19:30 dinner reservation). & $s_i \in \mathcal{R}_i$ \\
Operational Availability Window & The realized visit interval must lie within the admissible on-site window specified by the environment, including daily closure gaps and date-specific closures when applicable (e.g., restaurants closed 15:00--18:00 between lunch and dinner, museum closed on Mondays). & $[s_i, f_i] \subseteq \mathcal{W}_i$ \\
Minimum Dwell Time & Each visit must last at least a minimum duration determined by the attraction type and traveler profile (e.g., a deep museum visit for children requires $\geq$180 min, elderly visitors who skip stair-climbing areas need less time). & $f_i - s_i \geq p_i$ \\
Inter-Activity Feasibility & For consecutive activities $(i,j)$, the next activity must start after the previous activity finishes plus inter-location travel time and an arrival buffer. The agent must arrive at least $\delta_j$ minutes before the scheduled start time for security screening, ticket verification, or queuing. & $s_j \ge f_i + \tau_{ij} + \delta_j$ \\
\midrule
\rowcolor{headercolor}
\multicolumn{3}{@{}l}{\textbf{Cost Consistency Constraints} (Soft Constraints $\mathcal{C}_S$)} \\
Combinatorial Cost Matching & The agent must select appropriate ticket types or menu items (e.g., family pass vs. individual tickets, tasting menu vs. \`a la carte) and compute the total price based on official listed prices. & $C_i(y) = C_i^*$ \\
\midrule
\rowcolor{headercolor}
\multicolumn{3}{@{}l}{\textbf{Preference Constraints} (Soft Constraints $\mathcal{C}_S$)} \\
Lodging (Hotel) Selection & The agent must select the correct hotel from multiple candidates based on user preferences. The user may specify the hotel by name directly, or provide filtering criteria (e.g., city-center location) where only one option satisfies all requirements. & $v_{\text{hotel}} = v_{\text{hotel}}^*$ \\
\bottomrule
\end{tabular}%
}
\end{table*}

This section introduces WorldTravel, a benchmark for evaluating agents on real-world itinerary planning.
We formalize the task as a constrained scheduling problem in which every decision, from attraction entry time selection to restaurant reservation and hotel booking, is interdependent and governed by rigid temporal rules.
To ensure verifiability without sacrificing task complexity, we construct \textbf{WorldTravel-Webscape}, a high-fidelity web environment where agents must resolve information fragmentation by extracting constraints directly from visual interfaces. We validate its authenticity through a three-stage pipeline covering data curation, environment synthesis, and rigorous quality control.

\subsection{Task Formulation}

To ensure reliable evaluation, we abstract itinerary planning into a verifiable \textbf{constrained scheduling problem}. This approach strips away secondary real-world noise (e.g., stochastic delays) while preserving the \textit{minimal viable complexity} required to challenge an agent: \textbf{temporal coupling} and \textbf{information fragmentation}.

Given a user query $Q$, the agent must output an itinerary $y = (\mathcal{I}, \{s_i, d_i\}, \mathbf{v})$, comprising an ordered set of activities $\mathcal{I}$, their corresponding start times and durations $\{s_i, d_i\}$, and discrete decisions $\mathbf{v}$ (e.g., specific hotels or ticket selections). Feasibility is governed by parameters $\theta$ (e.g., reservation slots, operating windows, and travel buffers) that agents must extract directly from fragmented visual interfaces.

These decisions are \textbf{tightly coupled}: a single commitment, such as a 10:30 museum entry, propagates to restrict all downstream timing, often compressing the solution space to a unique valid itinerary. We quantify this complexity by the number of \textbf{temporal anchors} (fixed timed-entry activities), with tasks containing 0 to 5 anchors, spanning loosely to tightly coupled scenarios.

\subsection{Constraint Taxonomy}
\label{sec:constraints}
To balance real-world fidelity with unambiguous evaluation, we explicitly separate constraints that determine temporal feasibility from those that assess decision correctness.
Accordingly, we organize constraints into hard constraints $\mathcal{C}_H$
and soft constraints $\mathcal{C}_S$, as summarized in Table~\ref{tab:constraints}.

\textbf{Hard constraints} ($\mathcal{C}_H$) determine temporal feasibility and originate from real-world venue operations. They specify when and how long activities can occur (e.g., fixed entry slots for museums, minimum required visit durations, and mandatory arrival buffers).

\textbf{Soft constraints} ($\mathcal{C}_S$) capture decision correctness within feasible solutions and verify discrete choices (e.g., exact cost computation from listed prices or selecting the correct hotel among multiple candidates).



\subsection{Environment Construction}

As illustrated in Figure~\ref{fig:pipeline}, we build the benchmark through a three-stage pipeline grounded in real-world constraints.

\textbf{Stage 1: Data Collection.}
Real travel planning requires reconciling official policies with operational realities.
We collect \textit{factual data} from official venue websites (operating hours with mid-day closures, tiered pricing, timed-entry slots) and \textit{practical data} from travel platforms and user reviews (recommended visit durations, arrival buffers for security screening).
Both sources are necessary because official hours alone do not account for factors like 30-minute security queues at popular museums. Detailed collection specifications appear in Appendix~\ref{app:data_collection}.

\textbf{Stage 2: Webpage and Task Synthesis.}
We author static HTML pages using LLMs with manual review, introducing visual and structural diversity across pages to prevent template overfitting.
Information about each entity is distributed across multiple pages, forcing agents to integrate constraints from booking portals, travel guides, and menus.
Each city targets distinct reasoning challenges. For example, Rome requires choosing between skip-the-line and regular admission with different waiting times, while Florence includes venues with many stairs unsuitable for elderly visitors. City-specific task designs appear in Appendix~\ref{app:city-design}.

\textbf{Stage 3: Quality Control.}
To ensure tasks are neither trivial nor unsolvable, we adjust ticket availability and block specific time slots to create tightly coupled constraints, often reducing valid itineraries to a unique solution.
Each task undergoes three-round verification: author self-review, independent cross-solving, and external validation.
Pilot experiments on GPT-4.1, Doubao-1.8-Pro, and Gemini-2.5-Pro filter out edge cases.
This yields 150 tasks where real-world complexity is preserved while solutions remain automatically verifiable. Annotation protocols appear in Appendix~\ref{app:quality-control}.

\textbf{Agent Interface.}
We define 8 APIs covering attraction booking, restaurant reservations, hotel search, and route planning.
Unlike prior benchmarks that return structured JSON data, all APIs in WorldTravel return rendered webpage screenshots.
This design mirrors how human travelers interact with real booking platforms and prevents agents from exploiting fixed schemas.
Agents must locate relevant information within complex UI layouts, distinguish available from sold-out time slots via visual cues, and cross-reference data across multiple pages to recover complete constraints. API specifications appear in Appendix~\ref{sec:appendix_apis}.

\subsection{Evaluation Protocol}
\label{sec:evaluation}

WorldTravel employs fully automated evaluation through constraint verification functions.
Each task is annotated with constraints $\mathcal{C} = \mathcal{C}_H \cup \mathcal{C}_S$ and verification functions $v_c: \mathcal{Y} \to \{0, 1\}$ that return 1 if constraint $c$ is satisfied.
Detailed verification specifications appear in Appendix~\ref{app:verification}.

We measure performance through three metrics that distinguish temporal feasibility from decision quality, where $\mathcal{F}$ denotes the set of tasks whose itineraries satisfy all hard constraints.

\begin{table}[t]
\centering
\small
\caption{\textbf{Comparison with related benchmarks.} WorldTravel uniquely combines screenshot-based interaction with tightly coupled constraint reasoning (15+ constraints per task).}
\label{tab:benchmark_comparison}
\begin{tabular*}{\columnwidth}{@{\extracolsep{\fill}}llccc@{}}
\toprule
\textbf{Benchmark} & \textbf{Category} & \textbf{\# Tasks} & \textbf{Input} & \textbf{Avg. $|\mathcal{C}|$} \\
\midrule
TravelPlanner & Travel Planning & 1,225 & Text & $\sim$3 \\
ChinaTravel & Travel Planning & 1,154 & Text & $\sim$8 \\
TripTailor & Travel Planning & $\sim$4K & Text & $\sim$6 \\
\midrule
OSWorld & Interactive Agent & 369 & Screenshots & -- \\
WebArena & Interactive Agent & 812 & Screenshots & -- \\
TheAgentCompany & Interactive Agent & $\sim$175 & Screenshots & -- \\
\midrule
\textbf{WorldTravel} & \textbf{Both} & \textbf{150} & \textbf{Screenshots} & \textbf{15+} \\
\bottomrule
\end{tabular*}
\end{table}

\textbf{Feasibility Rate.}
The proportion of tasks yielding feasible itineraries is defined as
\begin{equation*}
\text{Feasibility Rate} = \frac{|\mathcal{F}|}{N}
\end{equation*}
This metric measures the ability to construct temporally coherent schedules under tightly coupled constraints, where a single scheduling error can cascade to render the entire itinerary infeasible.

\textbf{Constraint Violation.}
The average hard constraint violation rate is defined as
\begin{equation*}
\text{Constraint Violation} = 1 - \frac{1}{N} \sum_{i=1}^{N} \frac{n_H^{(i)}}{|\mathcal{C}_H^{(i)}|}
\end{equation*}
where $n_H^{(i)}$ is the number of satisfied hard constraints for task $i$.
This metric diagnoses partial constraint satisfaction.

\textbf{Optimality $\mid$ Feasible.}
The average soft constraint satisfaction among feasible solutions is defined as
\begin{equation*}
\text{Optimality} \mid \text{Feasible} = \frac{1}{|\mathcal{F}|} \sum_{i \in \mathcal{F}} \frac{n_S^{(i)}}{|\mathcal{C}_S^{(i)}|}
\end{equation*}
where $n_S^{(i)}$ is the number of satisfied soft constraints for task $i$.
This metric measures discrete decision accuracy within feasible solutions, testing whether agents correctly compute costs and select entities matching user preferences.

\subsection{Benchmark Statistics}
\label{sec:statistics}

WorldTravel comprises 150 tasks across 5 European cities (Berlin, Vienna, Rome, Barcelona, Florence), with 30 tasks per city.
WorldTravel-Webscape includes 36 attractions, 25 restaurants, and 26 hotels, supported by over 2,000 webpages covering booking interfaces, travel guides, menus, and transportation matrices.
Each task involves between 10 and 23 tightly coupled constraints, averaging 15+ per task.
Tasks range from 0 to 5 temporal anchors in count, with the majority concentrated at 2 or 3 anchors.
Detailed distributions appear in Appendix~\ref{app:statistics}.

Table~\ref{tab:benchmark_comparison} compares WorldTravel with related benchmarks.
Existing benchmarks either provide structured text with loosely coupled constraints (travel planning) or use screenshots without dense constraint reasoning (interactive agents).
WorldTravel combines screenshot-based parameter extraction with coordination of 15+ tightly coupled constraints.

\section{Experiments}
\label{sec:experiments}

\subsection{Experimental Setup}
\label{subsec:exp-setup}

\textbf{Model Configuration.}
We evaluate a range of large language models (LLMs) and vision-language models (VLMs), including both proprietary and open-source systems. For LLMs, we evaluate GPT-5.2~\citep{gpt5.2}, GPT-5.1~\citep{gpt5.1}, Deepseek v3.2~\citep{deepseekv3.2}, Claude-Opus-4.5~\citep{claude-opus-4.5}, Claude-Sonnet-4.5~\citep{claude-sonnet-4.5}, Gemini-3-Pro~\citep{gemini3pro}, Gemini-2.5-Pro~\citep{gemini2.5pro}, Doubao-1.8-Pro~\citep{bytedance_seed1.8_2025}, GLM-4.6~\citep{zaiorg_glm4.6_2025}, and Qwen-3-A235B~\citep{qwen_qwen3-235b-a22b-instruct-2507_2025}. For VLMs, we evaluate GPT-5.2, GPT-5.1, Claude-Opus-4.5, Claude-Sonnet-4.5, Gemini-3-Pro, Gemini-2.5-Pro, and Doubao-1.8-Pro. For proprietary models, we use the official inference APIs with default settings. For open-source models, we adopt a unified decoding configuration with temperature set to 1.0 and top-p to 0.7, while all other hyperparameters follow their respective defaults.

\textbf{Benchmark Configuration.}
Agents interact with the environment through API calls to retrieve constraint parameters. We evaluate models in two settings based on the API return format: in the text setting, APIs return structured textual descriptions; in the multi-modal setting, APIs return rendered webpage screenshots.

\textbf{Evaluation Metrics.}
We follow Section~\ref{sec:evaluation} and report three metrics: Feasibility Rate, Constraint Violation, and Optimality $\mid$ Feasible. 
To establish task difficulty, we conduct human evaluation with 5 participants (all with graduate-level education).

\begin{table}[!t]
\centering
\caption{\textbf{Main results on WorldTravel.} We evaluate 10 LLMs and 7 VLMs under text and multi-modal settings. Human performance is measured on a sample of 30 tasks.}
\label{tab:main_results}
\renewcommand{\arraystretch}{1.1}
\begin{tabularx}{\columnwidth}{@{}l >{\centering\arraybackslash}X >{\centering\arraybackslash}X >{\centering\arraybackslash}X@{}}
\toprule
\textbf{Model} & \textbf{Feasibility Rate (\%)}\,$\uparrow$ & \textbf{Constraint Violation (\%)}\,$\downarrow$ & \textbf{Optimality $|$ Feasible (\%)}\,$\uparrow$ \\
\midrule
\rowcolor{blue!7}
\textbf{Human Performance} & \textbf{77.86} & \textbf{6.48} & \textbf{84.76} \\
\midrule
\rowcolor{gray!6}
\multicolumn{4}{l}{\textit{Text Setting}} \\
GPT-5.2         & \first{32.67} & 21.92 & 85.34 \\
Claude-Opus-4.5 & \second{21.33} & 25.76 & 73.22 \\
GPT-5.1           & 16.67 & 32.85 & 83.80 \\
Gemini-3-Pro      & 14.67 & 32.87 & 77.34 \\
Deepseek v3.2     & 14.00 & 34.16 & 70.63 \\
Claude-Sonnet-4.5   & 10.00 & 34.43 & 59.77 \\
Doubao-1.8-Pro    & 8.67 & 37.79 & 77.18 \\
Gemini-2.5-Pro    & 8.00 & 41.14 & 83.19 \\
GLM-4.6           & 8.00 & 43.44 & 64.16 \\
Qwen-3-A235B      & 2.67 & 50.47 & 60.00 \\
\midrule
\rowcolor{gray!6}
\multicolumn{4}{l}{\textit{Multi-modal Setting}} \\
GPT-5.2           & \first{19.33} & 33.18 & 76.83 \\
Claude-Opus-4.5   & \second{14.00} & 30.86 & 77.61 \\
Gemini-3-Pro      & 8.00 & 40.10 & 80.41 \\
Doubao-1.8-Pro    & 4.00 & 46.63 & 63.33 \\
GPT-5.1           & 2.67 & 47.98 & 75.00 \\
Claude-Sonnet-4.5 & 2.67 & 44.51 & 55.00 \\
Gemini-2.5-Pro    & 1.33 & 52.71 & 60.00 \\
\bottomrule
\end{tabularx}
\end{table}

\subsection{Main Results}
We evaluate 10 LLMs under the text setting and 7 VLMs under the multi-modal setting. Table~\ref{tab:main_results} presents the results, from which we derive three key findings.

\textbf{Temporal constraint satisfaction is the core bottleneck for current models.}
The best-performing model GPT-5.2 achieves a Feasible Rate of only 32.67\% in the text setting, indicating that a substantial majority of tasks fail to produce executable itineraries. WorldTravel's hard constraints are tightly coupled: a single scheduling decision propagates constraints to all subsequent activities. For instance, booking a 10:30 museum entry forces lunch to be delayed past 12:00 and compresses the afternoon time window. Models must simultaneously coordinate multiple temporal anchors to construct feasible itineraries, rather than making locally optimal decisions in isolation. This result indicates a systematic deficiency in temporal constraint satisfaction, rather than simple information extraction or instruction-following failures.

\textbf{Proprietary models significantly outperform open-source models.}
Proprietary models (GPT-5.2, GPT-5.1, Claude-Opus-4.5) achieve Feasible Rates above 15\%, while open-source models (GLM-4.6, Qwen-3-A235B) remain below 10\%, with Qwen-3-A235B at only 2.67\%. Constraint Violation exhibits the same trend: GPT-5.2 violates 21.92\% of hard constraints on average, whereas Qwen-3-A235B violates 50.47\%. This suggests that temporal constraint satisfaction capability is highly correlated with model scale and training resources.

\textbf{Extracting constraint parameters from webpage screenshots constitutes an independent bottleneck.}
GPT-5.2 drops from 32.67\% in the text setting to 19.33\% in the multi-modal setting (a 13.34 pp decline), while Claude-Opus-4.5 declines from only 21.33\% to 14.00\% (a 7.33 pp decline). The multi-modal setting requires agents to extract constraint parameters from rendered webpage screenshots, including distinguishing sold-out from available time slots and integrating information scattered across multiple pages. This differentiated decline pattern indicates that extracting structured information from visual interfaces is a capability dimension largely independent of text-based reasoning: strong text reasoning ability does not imply accurate constraint extraction from screenshots.
\section{Discussion}
\label{sec:discussion}
\begin{table*}[!t]
\centering
\begin{minipage}{0.48\textwidth}
  \centering
  \caption{\textbf{Feasibility-Only ablation.} Removing soft constraints on the full 150-task benchmark yields modest Feasibility Rate improvements.}
  \label{tab:feasibility_ablation}
  \resizebox{0.9\linewidth}{!}{%
  \begin{tabular}{@{}l ccc@{}}
  \toprule
  \textbf{Model} & \textbf{Standard (\%)} & \textbf{F-Only (\%)} & \textbf{$\Delta$ (pp)} \\
  \midrule
  \rowcolor{gray!6}
  \multicolumn{4}{l}{\textit{Text Setting}} \\
  GPT-5.2 & 32.67 & 33.33 & +0.66 \\
  GPT-5.1 & 16.7 & 20.00 & +3.3 \\
  Gemini-3-Pro & 14.67 & 17.33 & +2.66 \\
  Doubao-1.8-Pro & 8.67 & 18.00 & +9.33 \\
  \midrule
  \rowcolor{gray!6}
  \multicolumn{4}{l}{\textit{Multi-modal Setting}} \\
  GPT-5.2 & 19.33 & 21.33 & +2.0 \\
  GPT-5.1 & 2.67 & 8.67 & +6.0 \\
  Gemini-3-Pro & 8.00 & 6.00 & $-$2.0 \\
  Doubao-1.8-Pro & 4.00 & 10.00 & +6.0 \\
  \bottomrule
  \end{tabular}%
  }
\end{minipage}
\hfill
\begin{minipage}{0.48\textwidth}
  \centering
  \caption{\textbf{Gold-$\theta$ ablation results.} Providing all hard constraint parameters in structured form on 30 high-coupling tasks. Substantial failures persist even with perfect parameter access.}
  \label{tab:gold_theta}
  \resizebox{\linewidth}{!}{%
  \begin{tabular}{@{}l ccc ccc@{}}
  \toprule
  & \multicolumn{3}{c}{\textbf{Feasibility Rate (\%)}} & \multicolumn{3}{c}{\textbf{Violation (\%)}} \\
  \cmidrule(lr){2-4} \cmidrule(lr){5-7}
  \textbf{Model} & \textbf{Std} & \textbf{Gold-$\theta$} & \textbf{$\Delta$} & \textbf{Std} & \textbf{Gold-$\theta$} & \textbf{$\Delta$} \\
  \midrule
  GPT-5.2 & 36.67 & 36.67 & +0 & 18.16 & 20.64 & +2.48 \\
  GPT-5.1 & 13.33 & 16.67 & +3.34 & 32.28 & 29.40 & $-$2.88 \\
  Deepseek v3.2 & 10.00 & 10.00 & +0 & 38.62 & 34.32 & $-$4.3 \\
  Gemini-3-Pro & 16.67 & 20.00 & +3.33 & 38.94 & 33.74 & $-$5.2 \\
  Doubao-1.8-Pro & 13.33 & 3.33 & $-$10 & 37.16 & 33.13 & $-$4.03 \\
  \bottomrule
  \end{tabular}%
  }
\end{minipage}
\end{table*}

We address three questions:
(1) \textbf{Is} planning itself a bottleneck, independent of parameter access?
(2) \textbf{How} much do soft constraints impair hard-constraint satisfaction?
(3) \textbf{Where} do current models reach their capacity limits?

\subsection{Planning Remains a Bottleneck Under Perfect Parameter Access}

In the standard setup, models must complete two stages. First, they retrieve constraint parameters via API interactions. Second, they plan the itinerary based on these parameters. To probe whether improved parameter access alone resolves failures, we design the \textbf{Gold-$\theta$ ablation}, where all hard constraint parameters $\theta$ (i.e., appointment slots, operational windows, minimum dwell times, transit times, and arrival buffers) are provided directly in a structured format, bypassing parameter acquisition entirely. We provide this structured information once in the system prompt, see in Appendix~\ref{sec:appendix_prompts_ablation} for details.

We select 30 tasks with the highest temporal coupling from the 150-task benchmark. These tasks contain the maximum number (5) of temporal anchors, which cannot be directly extracted from webpages but require joint reasoning over multiple booking systems and itinerary structures.

As shown in Table~\ref{tab:gold_theta}, providing Gold-$\theta$ parameters does not improve the Feasibility Rate of GPT-5.2 from 36.67\% to 36.67\% (+0 pp), while GPT-5.1 shows a similarly limited improvement, improving only marginally from 13.33\% to 16.67\% (+3.34 pp). However, even under ideal conditions with all parameters explicitly provided, GPT-5.2 achieves a Feasibility Rate of only 36.67\%, suggesting that \textbf{improving parameter access alone is insufficient} and that substantial errors persist under tightly coupled temporal constraints.

\subsection{Soft Constraints Have Limited Impact on Hard Constraint Satisfaction}

Under the standard setting, models must satisfy both hard constraints (temporal feasibility) and soft constraints (price calculation and hotel selection). To test whether this multi-objective load degrades hard constraint satisfaction, we conduct a \textbf{Feasibility-Only ablation}, where the system prompt instructs the model to focus exclusively on temporal feasibility while ignoring soft constraints (See Appendix~\ref{sec:appendix_prompts_ablation}). This experiment runs on the full 150-task benchmark.

\begin{figure}[t]
\centering
\begin{minipage}{0.48\columnwidth}
  \centering
  \includegraphics[width=\linewidth]{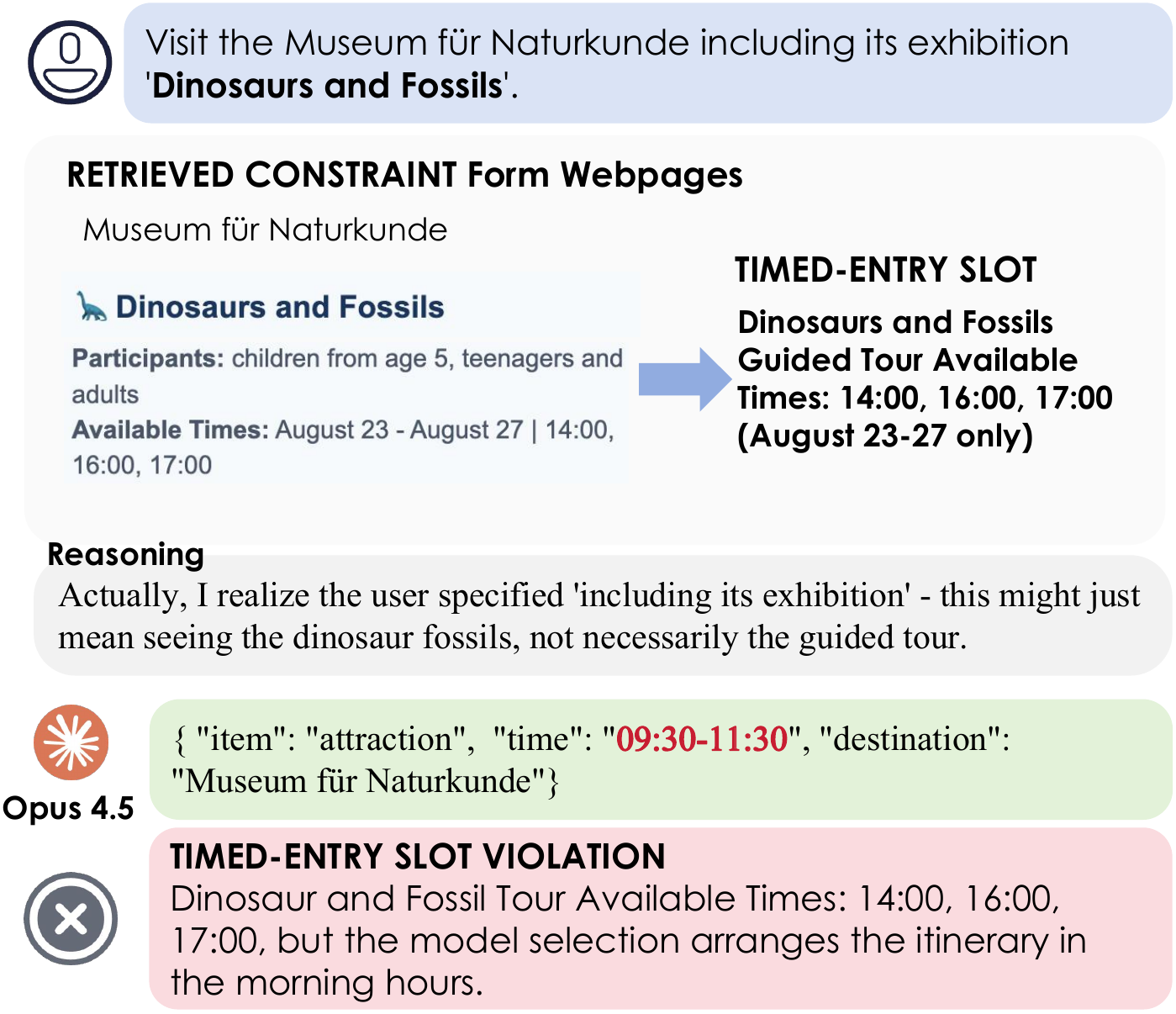}
  \caption{\textbf{Timed-Entry Slot violation.} The exhibition tour is available at 14:00, 16:00, and 17:00, but the model schedules the visit in the morning.}
  \label{fig:case_slot}
\end{minipage}
\hfill
\begin{minipage}{0.48\columnwidth}
  \centering
  \includegraphics[width=\linewidth]{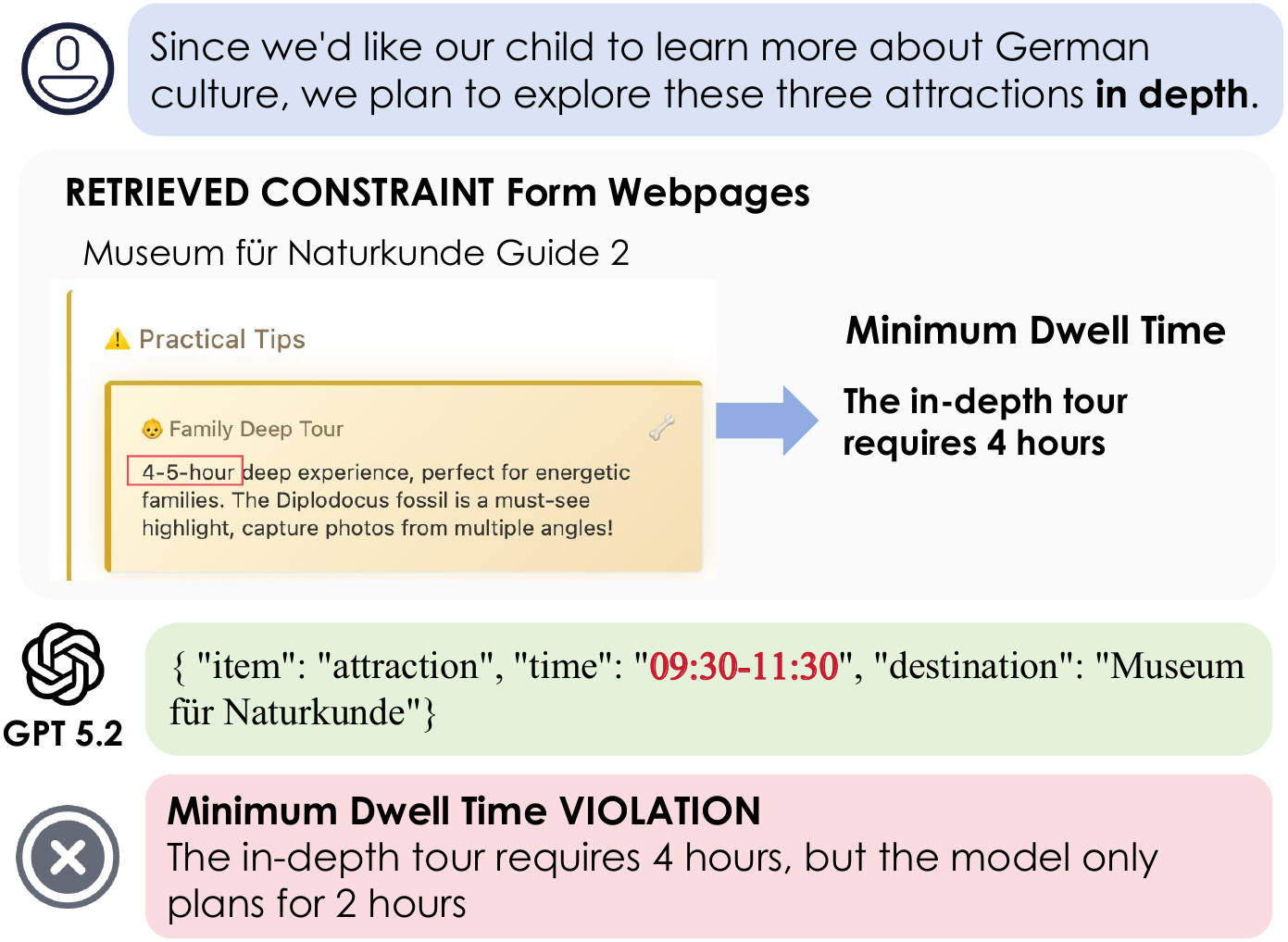}
  \caption{\textbf{Minimum Dwell Time violation.} The user requests an in-depth tour (4--5 hours required), but the model schedules only 2 hours.}
  \label{fig:case_dwell}
\end{minipage}
\end{figure}

As Table~\ref{tab:feasibility_ablation} illustrates, removing soft constraints leads to consistent improvements across both settings, with GPT-5.2 improving from 32.67\% to 33.33\% (+0.66 pp) in the text setting and from 19.33\% to 21.33\% (+2.0 pp) in the multi-modal setting. The slightly larger gain in the multi-modal setting likely reflects the higher cognitive load from simultaneous visual extraction and constraint reasoning.

Nevertheless, even in this feasibility-only setting, the Feasibility Rate remains low for GPT-5.2 at 33.33\% in Text and 21.33\% in Multi-modal, indicating that \textbf{soft constraints have limited impact on hard constraint satisfaction} and that the core difficulty lies in temporal reasoning.

\subsection{Planning Capability Exhibits a Clear Capacity Limit}

Given that planning reasoning is the core bottleneck, we further characterize its failure modes along two dimensions: constraint type and constraint quantity.

\textbf{Timed-Entry Slots are the Most Challenging Constraint Type.} We analyze satisfaction rates for three hard constraint categories: Minimum Dwell Time, Timed-Entry Slot \& Operational Availability Window, and Inter-Activity Feasibility. In the text setting, these constraints achieve average satisfaction rates of 42\%, 27\%, and 40\%, respectively. The multi-modal setting shows uniformly lower performance: 30\% ($-$12 pp), 16\% ($-$11 pp), and 35\% ($-$5 pp).

Timed-Entry Slots consistently exhibit the lowest satisfaction rate in both settings. In particular, they achieve only 27\% in the text setting and 16\% in the multi-modal setting, making them the most challenging constraint type in absolute terms. This pattern holds across individual models (see Appendix~\ref{sec:appendix_constraint}). The difficulty likely stems from the requirement for precise temporal alignment: the model must select a specific slot from discrete options while ensuring compatibility with all downstream activities. As illustrated in Figures~\ref{fig:case_dwell} and~\ref{fig:case_slot}, typical failures include allocating insufficient dwell time and selecting unavailable entry slots.

\begin{figure}[!t]
\centering
\begin{subfigure}[b]{0.48\textwidth}
\centering
\includegraphics[width=\textwidth]{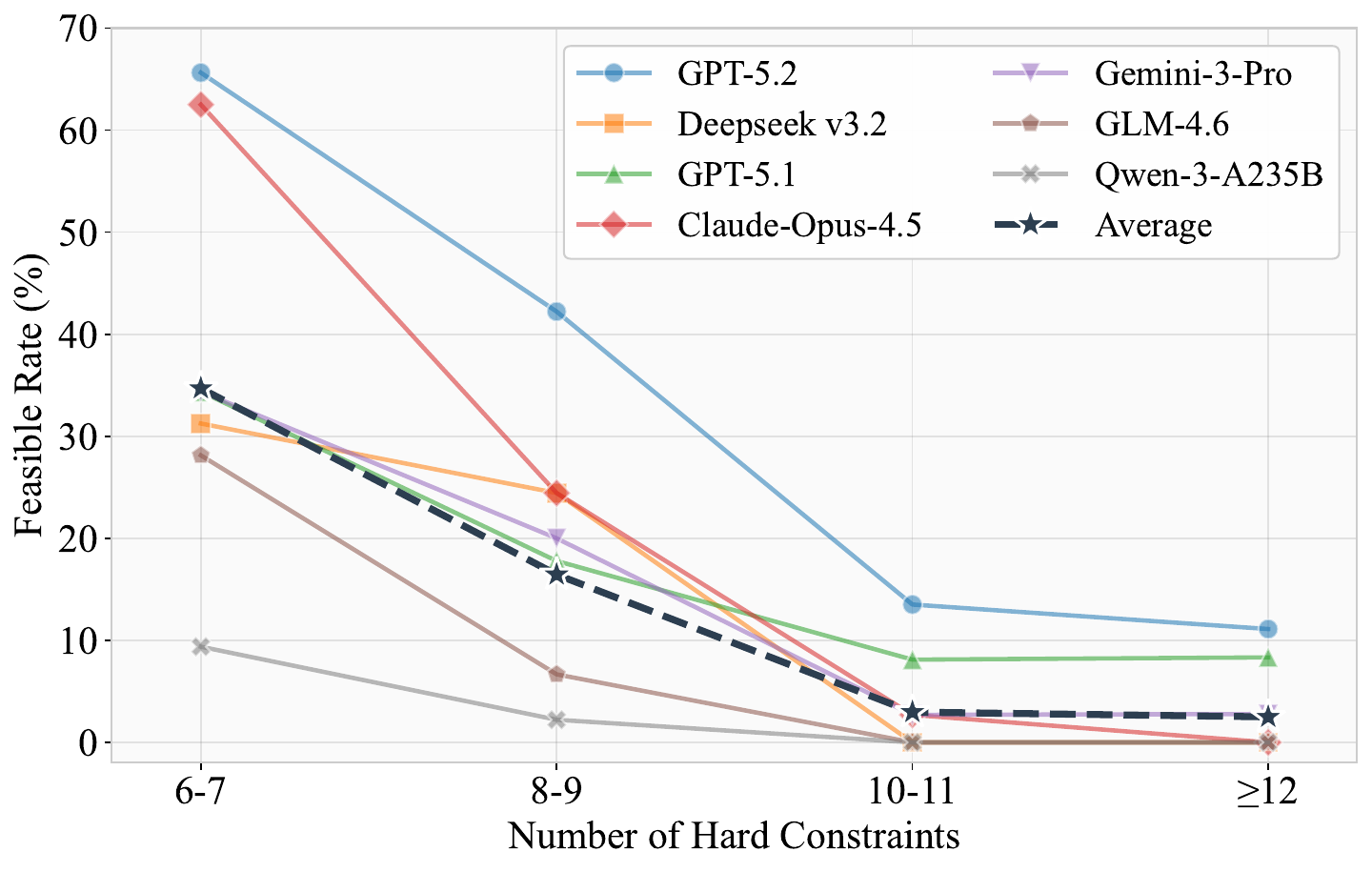}
\caption{Feasibility rate vs. constraint count (text setting).}
\end{subfigure}
\hfill
\begin{subfigure}[b]{0.48\textwidth}
\centering
\includegraphics[width=\textwidth]{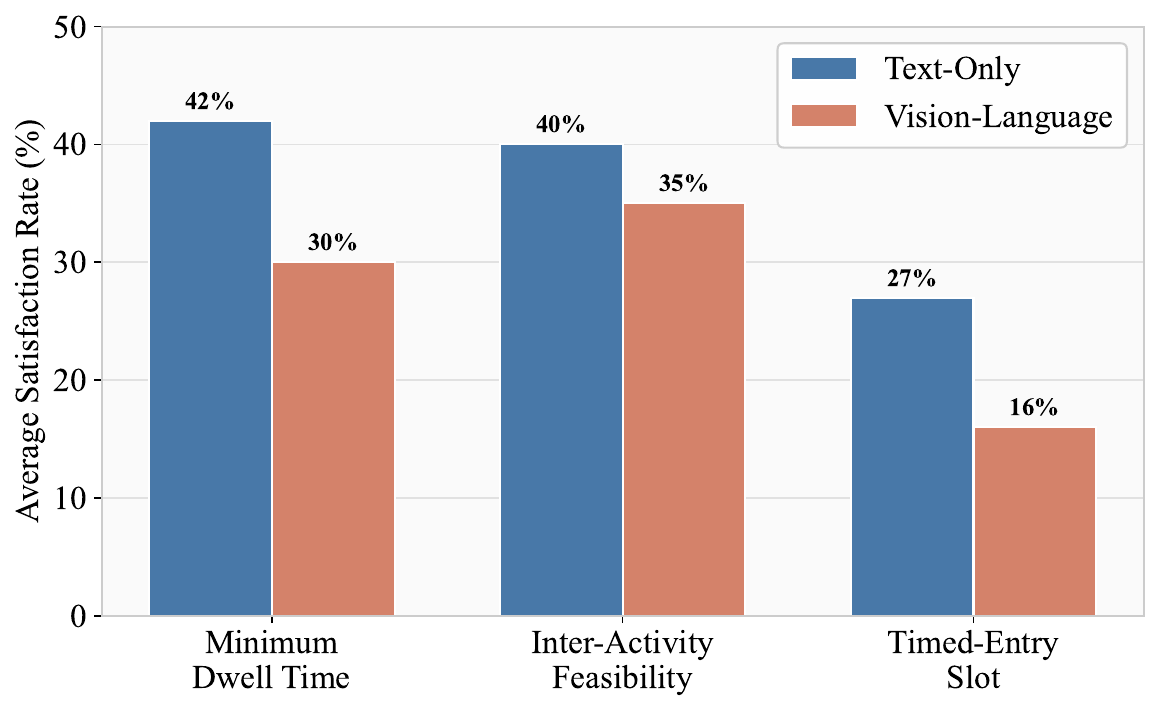}
\caption{Satisfaction rate by constraint type (text vs. vision).}
\end{subfigure}
\caption{\textbf{Constraint analysis.} (a) Feasibility collapses when constraints exceed 10, even when constraint parameters are provided as structured text. (b) Timed-Entry Slots show the lowest satisfaction rates across all models, with vision-based extraction further degrading performance.}
\label{fig:constraint_analysis}
\end{figure}

\textbf{Feasibility Rate Collapses Beyond 10 Constraints.} We group tasks by the number of hard constraints and calculate the Feasibility Rate for each group. As illustrated in Figure~\ref{fig:constraint_analysis}(left), the average Feasibility Rate declines from 36.3\% at 6--7 constraints to 18.3\% at 8--9 constraints, then collapses to 3.4\% at 10--11 constraints and 3.1\% beyond 12 constraints.

This pattern reveals a distinct capacity threshold at approximately 10 constraints. GPT-5.2 exhibits the greatest resilience, maintaining over 40\% feasibility up to 9 constraints and still achieving 13.5\% feasibility at 10--11 constraints. In contrast, weaker models collapse much earlier: GLM-4.6 drops from 28.1\% to 0\% as constraints increase from 6--7 to 10--11, and Qwen-3-A235B follows a similar trajectory (9.4\% $\rightarrow$ 0\%).

These findings reveal \textbf{where current models reach their capacity limits}.
A threshold emerges around 10 constraints, beyond which feasibility collapses across model families.
Timed-entry slots are the hardest constraint type.
The multi-modal setting compounds this difficulty, as models must extract parameters from screenshots while reasoning about temporal dependencies.

\section{Conclusion}
\label{sec:conclusion}

We introduce WorldTravel, a benchmark for evaluating LLMs on constraint-intensive travel itinerary planning. WorldTravel features 150 tasks with an average of 15+ tightly coupled temporal constraints, where agents must extract parameters from rendered webpage screenshots and coordinate scheduling decisions that propagate across the entire itinerary.

Evaluation of 10 frontier models reveals that global temporal reasoning remains a fundamental challenge: the best model achieves only 32.67\% Feasibility Rate in the text setting, dropping to 19.33\% in the multi-modal setting. 
Ablation studies show that planning reasoning itself is the primary bottleneck---even with perfect parameter access, feasibility reaches only 36.67\% on high-coupling tasks---and identify a capacity limit at approximately 10 constraints, beyond which feasibility collapses. These findings suggest that perception and reasoning constitute two independent bottlenecks, both essential for real-world planning tasks.

\section{Contributions}
\textbf{Project Lead}\\
Zexuan Wang$^*$, 
Chenghao Yang$^*$, 
Yingqi Que\\
(\email{zexuan.jason@bytedance.com, yangchenghao17@gmail.com})\\

\textbf{Core Contributors}\\
Zhenzhu Yang$^\S$, 
Huaqing Yuan$^\S$, 
Yiwen Wang$^\S$, 
Zhengxuan Jiang$^\S$, 
Shengjie Fang$^\S$, 
Zhenhe Wu$^\S$, 
Zhaohui Wang$^\S$, 
Zhixin Yao\\

\textbf{Contributors}\\
Jiashuo Liu, Jincheng Ren$^\S$, Yuzhen Li, Yang Yang, Jiaheng Liu$^\S$, Jian Yang$^\S$ \\

\textbf{Advisors}\\
Zaiyuan Wang$^\ddagger$ \\
Ge Zhang \\
Zhoufutu Wen$^\dagger$ (\email{liniuniu@bytedance.com}) \\
Wenhao Huang (\email{huang.wenhao@bytedance.com})\\

$^*$ denotes equal contribution.\\
$^\S$ denotes affiliation with M-A-P.\\
$^\ddagger$ denotes that the author was previously affiliated with ByteDance Seed and later joined HUMANLAYA.\\
$^\dagger$  denotes the corresponding author.\\
Contributors without explicit affiliations are from ByteDance Seed.
During this work, Chenghao Yang is an intern at ByteDance Seed.\\

\clearpage

\bibliographystyle{plainnat}
\bibliography{main}

\clearpage

\beginappendix

\section{Data Collection Specifications}
\label{app:data_collection}

This appendix details the data collection process that grounds WorldTravel in real-world travel planning scenarios. Annotators collect two categories of data from different sources to construct the constraint parameters $\theta$ and task content.

\subsection{Factual Data from Official Sources}

Factual data provides authoritative constraint parameters extracted from official venue websites and booking platforms. This category includes:

\begin{itemize}
    \item \textbf{Operating Hours:} Daily opening and closing times for attractions and restaurants, including mid-day closure gaps (e.g., restaurants closed between 15:00 and 18:00 for lunch-dinner transitions) and date-specific closures (e.g., museums closed on Mondays).

    \item \textbf{Tiered Pricing Structures:} Official ticket prices with multiple fare categories (e.g., Adult, Child, Student, EU Citizen, Senior) for attractions, and menu pricing for restaurants.

    \item \textbf{Timed-Entry Slot Availability:} For reservation-required venues, the discrete time slots available for booking (e.g., 10:00, 10:30, 11:00 museum entry slots; 19:00, 19:30, 20:00 dinner reservations) along with availability status.

    \item \textbf{Ticket Variants and Add-ons:} Different ticket types (e.g., Skip-the-Line vs. Standard Admission) and optional services (e.g., Audio Guides, Guided Tours with language options) with corresponding prices.

    \item \textbf{Exhibition and Performance Schedules:} Temporary exhibitions, special events, and performance times for venues offering time-specific programming.

    \item \textbf{Hotel Information:} Nightly rates, room types, location features, and amenities from hotel booking sites.
\end{itemize}

\subsection{Practical Data from Travel Platforms}

Practical data captures operational patterns and user experiences extracted from travel review platforms (e.g., TripAdvisor), travel blogs, and user-generated content. This category includes:

\begin{itemize}
    \item \textbf{Recommended Visit Duration:} Suggested time allocations for attractions reflecting different visit styles (e.g., 90 minutes for a quick tour vs. 180 minutes for an in-depth museum visit), and typical dining durations for restaurants.

    \item \textbf{Arrival Buffers:} Practical timing recommendations for security screening, ticket verification, and queuing at popular venues (e.g., arrive 15 minutes early for timed-entry attractions).

    \item \textbf{Optimal Visiting Periods:} Crowd avoidance strategies, best times for photography, and seasonal considerations extracted from traveler experiences.

    \item \textbf{Logistical Details:} Boarding locations for river cruises, meeting points for walking tours, and other venue-specific operational information not always clear from official sources.

    \item \textbf{User Reviews:} Qualitative feedback covering service quality, food experiences, ambiance, and practical tips from verified travelers.
\end{itemize}

\subsection{Data Grounding and Validation}

All collected data undergoes validation to ensure consistency between factual and practical sources. For example, if a museum's official website lists 09:00-18:00 operating hours, but user reviews consistently mention 30-minute security lines in peak season, annotators record both the official window $\mathcal{W}_i = [09:00, 18:00]$ and the practical arrival buffer $\delta_i = 30$ minutes. This dual-source approach ensures that constraint parameters reflect both official policies and operational realities, preserving the authenticity of real-world travel planning.

\section{City-Specific Task Design}
\label{app:city-design}

WorldTravel currently includes five cities with distinct task designs.
Each city focuses on attraction-centered planning scenarios that emphasize temporal feasibility under diverse realistic constraints.
Table~\ref{tab:city-design-attraction} summarizes several representative task focuses for each city.

\begin{table}[H]
\centering
\caption{City task focuses in WorldTravel.}
\label{tab:city-design-attraction}
\renewcommand{\arraystretch}{1.2}
\begin{tabularx}{\textwidth}{p{2cm} X}
\toprule
\textbf{City} & \textbf{Task Focuses} \\
\midrule

\textbf{Rome} &
\textbf{Queue-aware temporal feasibility.} Certain attractions provide multiple admission modes (e.g., skip-the-line vs. regular), each with different prices, and required waiting times.

\textbf{Multi-attraction pass ticket.} Certain attractions are governed by bundled passes with explicit coverage and ordering constraints.

\textbf{Cross-page venue conflict.} Event information may be inconsistent across webpages, such as when a performance is associated with one institution but held at a different physical venue.

\textbf{Performance duration reasoning.} If no explicit duration is provided, the agent must first determine the type of performance (e.g., opera or ballet) and use it to estimate the show length for scheduling.

\textbf{Color-aware reasoning.} Inferring the mapping between time slots and supported languages from color-coded page elements. \\

\midrule

\textbf{Florence} &
\textbf{Multi-part attraction access and physical feasibility.} Large attraction complexes often include multiple sub-sites with different opening times and entry rules. Some areas also involve many stairs or steep climbs, which may not be suitable for all visitors.

\textbf{Artwork-focused route choice.} Some attractions provide multiple guided routes marked in different colors on floor maps, each passing through different exhibition areas. \\

\midrule

\textbf{Berlin} &
\textbf{Scheduling under limited tickets and exhibition constraints.} Ticket availability may vary by time slot, and some exhibitions are only accessible at specific times. The agent must calculate for remaining ticket numbers and exhibition schedules when choosing entry times.

\textbf{Stay time based on user interests.} Different attractions support both in-depth visits and quick overviews. \\

\midrule

\textbf{Vienna} &
\textbf{Daily planning around performances.} Different operas and ballets are offered on different dates and at fixed times, with varying durations and seat availability.

\textbf{Program filtering by language support.} Performances differ in spoken language, subtitle support, and whether pre-show explanations are provided. \\

\midrule

\textbf{Barcelona} &
\textbf{Ticket choice based on visit experience.} Attractions often offer multiple ticket types with different included areas, immersive features, or special experiences.

\textbf{Guided tours and language options.} Many attractions provide guided tours in multiple languages with limited time slots.

\textbf{Choosing visiting times for photography.} Some attractions recommend specific times for photography or viewing based on lighting conditions. \\

\bottomrule
\end{tabularx}
\end{table}

\section{Quality control}
\label{app:quality-control}

The quality control focuses on task validity, solution correctness, and evaluation reliability. It spans task construction, solution annotation, and multiple validation rounds.

\paragraph{Task construction.}
We adopt a hierarchical task construction strategy. 
Each high-coupling core scenario is first constructed as a feasible planning problem, and then expanded into multiple derived tasks by varying user profiles (e.g., language preferences, group size, age, or budget) and task settings (e.g., ticket types, menu options, or hotel preferences).

\paragraph{Solution annotation.}
All reference solutions are written manually by the task authors. 
Each solution is reviewed multiple times to remove ambiguity and to make sure that all constraints can be checked by automated evaluation functions.

During this process, real world travel planning challenges are deliberately converted into concrete reasoning targets. For example, places with many stairs and no elevators become unsuitable for seniors or children. In the end, these real world factors are expressed as time and cost constraints that can be checked automatically.

\paragraph{Multiple validation rounds.}
Each task goes through three rounds of verification. 
(i) The original author first reviews both the task and the solution.
(ii) A second annotator then solves the task independently and compares the result with the reference answer.
(iii) Finally, an external reviewer is hired to validate task clarity and answer correctness. Only tasks that pass all three rounds are included in the final benchmark.

Through this process, we curate a final set of 150 tasks across five European cities.

\noindent\textbf{Q: How do you ensure that newly created tasks do not affect the time settings of previously constructed ones?}

\noindent\textbf{A:} To avoid interference between tasks, we assign different date ranges to different task batches.

\vspace{0.5em}

\noindent\textbf{Q: How do you ensure that highly coupled tasks admit a unique feasible solution?}

\noindent\textbf{A:} To ensure solution uniqueness, we first build a valid reference itinerary and then modify the webpages accordingly, such as adjusting remaining ticket numbers, blocking certain time slots and visit durations. These changes enforce tightly coupled constraints and remove alternative feasible solutions.

\section{API Specifications}
\label{sec:appendix_apis}

This appendix provides a detailed specification of the APIs (tools) available to agents in the World Travel benchmark environment. 

\textbf{Key Design Choice.} Unlike prior benchmarks (e.g., TravelPlanner, ChinaTravel) that return structured JSON, all APIs in World Travel return \textbf{rendered webpage screenshots}. This design enforces multi-modal reasoning: agents must perform OCR, parse UI semantics (e.g., distinguish grayed-out unavailable slots from selectable ones), and integrate information across multiple tool calls---directly mirroring how human travelers interact with real booking platforms.

\subsection{Attraction-Related APIs}

\begin{apibox}{get\_attraction\_list}
\textbf{Description:} Returns a webpage screenshot \textbf{containing the complete list of attractions for any given city}. This page enumerates all attractions used in tasks and evaluation for that city, defining a \textbf{closed attraction set} from which agents must exclusively select when constructing itineraries. The page displays attraction cards with thumbnail images, attraction names, and brief descriptive summaries.

\vspace{0.8em}\noindent
\begin{minipage}[c]{0.18\linewidth}
\centering\textbf{Parameters:}
\end{minipage}%
\begin{minipage}[c]{0.82\linewidth}
\begin{tabularx}{\linewidth}{@{} l l X @{}}
\toprule
\textbf{Parameter} & \textbf{Type} & \textbf{Details} \\
\midrule
\texttt{city} & \textit{string} & \textbf{Required.} The name of the city. Must be selected from the predefined set of supported cities exposed in the system prompt (e.g., ``Berlin'').\\
\bottomrule
\end{tabularx}
\end{minipage}
\end{apibox}

\begin{apibox}{get\_attraction\_ticket\_availability}
\textbf{Description:} Returns a webpage screenshot of the booking portal for a \textbf{specific attraction} on a \textbf{specific date}. The page includes selectable and grayed-out unavailable time slots, event or show schedules, remaining admission capacity, and detailed ticket options with prices. Ticket options further specify visitor categories (e.g., individual, family, group) and add-on services such as guided tours and audio guides, both with available languages options.

\vspace{0.8em}\noindent
\begin{minipage}[c]{0.18\linewidth}
\centering\textbf{Parameters:}
\end{minipage}%
\begin{minipage}[c]{0.82\linewidth}
\begin{tabularx}{\linewidth}{@{} l l X @{}}
\toprule
\textbf{Parameter} & \textbf{Type} & \textbf{Details} \\
\midrule
\texttt{attraction} & \textit{string} & \textbf{Required.} The name of the attraction. Must be selected from the predefined attraction set exposed in the system prompt (e.g., ``Museum für Naturkunde''). \\
\addlinespace
\texttt{date} & \textit{string} & \textbf{Required.} The date of the query in ``Month.Day'' format (e.g., ``8.1''). Supported dates are restricted to August 1st through August 31st. \\
\bottomrule
\end{tabularx}
\end{minipage}
\end{apibox}

\begin{apibox}{get\_guide\_detail}
\textbf{Description:} Returns a webpage screenshot of user-generated travel guide content for a \textbf{specific attraction}. The guides provide recommended visit durations under distinct exploration modes (e.g., quick visits vs.\ in-depth tours), tailored suggestions for diverse visitor profiles (e.g., art enthusiasts), and practical constraints including physical demands, optimal visiting or photography times, and crowd-avoidance tips. These pages supply experiential and commonsense knowledge to support reasoning about visit duration, activity planning, and user intent alignment.

\vspace{0.8em}\noindent
\begin{minipage}[c]{0.18\linewidth}
\centering\textbf{Parameters:}
\end{minipage}%
\begin{minipage}[c]{0.82\linewidth}
\begin{tabularx}{\linewidth}{@{} l l X @{}}
\toprule
\textbf{Parameter} & \textbf{Type} & \textbf{Details} \\
\midrule
\texttt{attraction} & \textit{string} & \textbf{Required.} The name of the attraction. Must be selected from the predefined attraction set exposed in the system prompt (e.g., ``DDR Museum''). \\
\bottomrule
\end{tabularx}
\end{minipage}
\end{apibox}

\subsection{Restaurant-Related APIs}

\begin{apibox}{get\_restaurant\_list}
\textbf{Description:} Returns a webpage screenshot \textbf{containing the complete list of restaurants available for any given city}. This page enumerates all restaurants used in tasks and evaluation for that city, defining a \textbf{closed restaurant set} from which agents must exclusively select when constructing itineraries. The page displays restaurant cards with images, restaurant names, and brief descriptive summaries, along with high-level attributes such as dining style, cuisine tags, and quality indicators.

\vspace{0.8em}\noindent
\begin{minipage}[c]{0.18\linewidth}
\centering\textbf{Parameters:}
\end{minipage}%
\begin{minipage}[c]{0.82\linewidth}
\begin{tabularx}{\linewidth}{@{} l l X @{}}
\toprule
\textbf{Parameter} & \textbf{Type} & \textbf{Details} \\
\midrule
\texttt{city} & \textit{string} & \textbf{Required.} The name of the city. Must be selected from the predefined set of supported cities exposed in the system prompt (e.g., ``Berlin'').\\
\bottomrule
\end{tabularx}
\end{minipage}
\end{apibox}

\begin{apibox}{get\_restaurant\_reservation\_availability}
\textbf{Description:} Returns a webpage screenshot of the reservation portal for a \textbf{specific restaurant} on a \textbf{specific date}. The page includes selectable and grayed-out unavailable time slots, opening hours and dining sessions (e.g., lunch or dinner), and reservation availability information.

\vspace{0.8em}\noindent
\begin{minipage}[c]{0.18\linewidth}
\centering\textbf{Parameters:}
\end{minipage}%
\begin{minipage}[c]{0.82\linewidth}
\begin{tabularx}{\linewidth}{@{} l l X @{}}
\toprule
\textbf{Parameter} & \textbf{Type} & \textbf{Details} \\
\midrule
\texttt{restaurant} & \textit{string} & \textbf{Required.} The name of the restaurant. Must be selected from the predefined restaurant set exposed in the system prompt (e.g., “Rutz Restaurant”). \\
\addlinespace
\texttt{date} & \textit{string} & \textbf{Required.} The date of the query in ``Month.Day'' format (e.g., ``8.1''). Supported dates are restricted to August 1st through August 31st. \\
\bottomrule
\end{tabularx}
\end{minipage}
\end{apibox}

\begin{apibox}{get\_restaurant\_guide\_detail}
\textbf{Description:} Returns webpage screenshots of user-generated dining guide content for a \textbf{specific restaurant}, including experiential dining notes (e.g., recommended dining durations, and dish suggestions) and menu pages with dish listings and prices. These pages supply experiential, commonsense, and cost-related knowledge to support reasoning about dining duration, dish selection, meal planning, and alignment with user dining preferences and intent.

\vspace{0.8em}\noindent
\begin{minipage}[c]{0.18\linewidth}
\centering\textbf{Parameters:}
\end{minipage}%
\begin{minipage}[c]{0.82\linewidth}
\begin{tabularx}{\linewidth}{@{} l l X @{}}
\toprule
\textbf{Parameter} & \textbf{Type} & \textbf{Details} \\
\midrule
\texttt{restaurant} & \textit{string} & \textbf{Required.} The name of the restaurant. Must be selected from the predefined restaurant set exposed in the system prompt (e.g., “Facil”). \\
\bottomrule
\end{tabularx}
\end{minipage}
\end{apibox}

\subsection{Hotel-Related APIs}

\begin{apibox}{get\_hotel\_list}
\textbf{Description:} Returns a webpage screenshot \textbf{containing the complete list of hotels available for a given city}. This page enumerates all hotels used in tasks and evaluation for that city, defining a \textbf{closed hotel set} from which agents must exclusively select when constructing itineraries. The page presents hotel cards with images, hotel names, and brief descriptive summaries.

\vspace{0.8em}\noindent
\begin{minipage}[c]{0.18\linewidth}
\centering\textbf{Parameters:}
\end{minipage}%
\begin{minipage}[c]{0.82\linewidth}
\begin{tabularx}{\linewidth}{@{} l l X @{}}
\toprule
\textbf{Parameter} & \textbf{Type} & \textbf{Details} \\
\midrule
\texttt{city} & \textit{string} & \textbf{Required.} The name of the city. Must be selected from the predefined restaurant set exposed in the system prompt (e.g., “W Barcelona”). \\
\bottomrule
\end{tabularx}
\end{minipage}
\end{apibox}

\subsection{Transportation-Related APIs}

\begin{apibox}{get\_route\_info}
\textbf{Description:} 
Returns a webpage screenshot of the city-level \textbf{transportation matrix}. The page lists all supported locations and provides pairwise inter-location travel costs under multiple transportation modes (e.g., walking, driving, public transit, taxi).

\vspace{0.8em}\noindent
\begin{minipage}[c]{0.18\linewidth}
\centering\textbf{Parameters:}
\end{minipage}%
\begin{minipage}[c]{0.82\linewidth}
\begin{tabularx}{\linewidth}{@{} l l X @{}}
\toprule
\textbf{Parameter} & \textbf{Type} & \textbf{Details} \\
\midrule
\texttt{city} & \textit{string} & \textbf{Required.} The name of the city. Must be selected from the predefined set of supported cities exposed in the system prompt (e.g., ``Berlin''). \\
\bottomrule
\end{tabularx}
\end{minipage}
\end{apibox}

\subsection{Design Rationale}

The screenshot-based output design serves four purposes:

\begin{enumerate}
    \item \textbf{Realistic Evaluation.} It directly mirrors how human travelers interact with real booking platforms (e.g., TripAdvisor, Booking.com, Google Maps), ensuring that benchmark performance correlates with real-world deployment capability.
    
    \item \textbf{Multi-Modal Reasoning.} Agents must: (i) locate relevant information within complex UI layouts, (ii) distinguish available from unavailable options via visual cues (e.g., grayed-out buttons, sold-out labels), and (iii) cross-reference information across multiple tool calls to recover the complete constraint space.
    
    \item \textbf{Information Fragmentation.} By distributing constraint-relevant information across multiple page types (booking portals, guide pages, menus, transportation matrices), we enforce cross-page information aggregation---a capability essential for real-world agents but absent from prior benchmarks that pre-aggregate all information into structured returns.

    \item \textbf{Robustness to Format Overfitting.} By avoiding pre-structured returns, the environment prevents agents from exploiting fixed schemas or shortcut extraction patterns. Instead, agents must infer relevant fields from visual layouts and natural language descriptions, mitigating benchmark-specific overfitting and fosters transferable planning behaviors.
\end{enumerate}

\section{Verification Function Specifications}
\label{app:verification}

World Travel adopts a ``hard to solve, easy to verify'' evaluation paradigm. Each constraint maps to a verification function, enabling fully automated scoring without human judgment.

Given a generated itinerary $\mathcal{I}$, each constraint $c$ is evaluated by a boolean function:
\begin{equation*}
    \texttt{verify}_c(\mathcal{I}) =
    \begin{cases}
        1 & \text{if } \mathcal{I} \text{ satisfies } c, \\
        0 & \text{otherwise.}
    \end{cases}
\end{equation*}

We partition constraints into \textbf{hard constraints}, which determine itinerary feasibility, and \textbf{soft constraints}, which assess decision correctness within feasible plans.

\subsection{Hard Constraint Verification}

Hard constraints in WorldTravel enforce temporal feasibility of itineraries, where any violation directly renders an itinerary infeasible. These constraints mandate that agents perform joint reasoning over time-entry slot, minimum dwell time, and arrival buffers. Table~\ref{tab:hard-verification} summarizes the hard-constraint verification functions.

\begin{table}[h]
\centering
\caption{Hard constraint verification functions in WorldTravel. Each function returns a boolean value indicating whether the itinerary strictly satisfies the corresponding temporal feasibility requirement.}
\label{tab:hard-verification}
\resizebox{\textwidth}{!}{%
\begin{tabular}{@{}lll p{7.2cm}@{}}
\toprule
\textbf{Category} & \textbf{Function} & \textbf{Parameters} & \textbf{Verification Logic} \\
\midrule
\multirow{3}{*}{Temporal} 
& \texttt{if\_poi\_enough\_time()} 
& poi, date\_range, time\_constraint
& Returns \texttt{True} iff the total scheduled visit duration of \texttt{poi} within \texttt{date\_range} $\ge$ min\_duration. \\

& \texttt{if\_poi\_in\_time()} 
& poi, date\_range, target\_time\_range
& Returns \texttt{True} iff the scheduled entry time of \texttt{poi} within \texttt{date\_range} lies in \texttt{target\_time\_range}. \\

& \texttt{if\_poi\_start\_time\_delayed()} 
& poi, date\_range, delay\_minutes 
& Returns \texttt{True} iff the actual arrival time at \texttt{poi} satisfies arrival\_time $\le$ start\_time $-$ delay\_minutes. \\

\bottomrule
\end{tabular}%
}
\end{table}

\subsection{Soft Constraint Verification}

Soft constraints evaluate whether agents make correct and grounded decisions once temporal feasibility is satisfied. These constraints measure cost reasoning accuracy and hotel selection. Table~\ref{tab:soft-verification} summarizes the soft-constraint verification functions.

\begin{table}[h]
\centering
\caption{Soft constraint verification functions.}
\label{tab:soft-verification}
\resizebox{\textwidth}{!}{%
\begin{tabular}{@{}lll p{7.2cm}@{}}
\toprule
\textbf{Category} & \textbf{Function} & \textbf{Parameters} & \textbf{Verification Logic} \\
\midrule
Cost 
& \texttt{if\_poi\_cost\_matches()} 
& poi, item\_type, expected\_cost 
& Returns \texttt{True} iff the computed total cost of the specified \texttt{poi} (under the given \texttt{item\_type}) exactly matches \texttt{expected\_cost}. \\
\addlinespace
Preference 
& \texttt{if\_poi\_present()}
& poi, date\_range 
& Returns \texttt{True} iff the specified \texttt{poi} appears in the itinerary within \texttt{date\_range}. \\
\bottomrule
\end{tabular}%
}
\end{table}
\section{Benchmark Statistics}
\label{app:statistics}

This appendix provides detailed statistics on the World Travel benchmark, including environment scale and constraint distribution.

\subsection{Environment Scale}

Table~\ref{tab:env_scale} summarizes the environment scale across five European cities. The environment comprises a diverse set of real-world entities: attractions range from world-renowned museums and cathedrals to local markets and river cruises; restaurants span Michelin-starred fine dining to traditional local eateries; hotels cover luxury properties to boutique accommodations.

\begin{table}[h]
\centering
\caption{Environment scale by city. Webpages include daily booking interfaces (31 days per entity), travel guides, menus, hotel listings, and transportation matrices.}
\label{tab:env_scale}
\small
\begin{tabular}{@{}lcccccc@{}}
\toprule
& \textbf{Berlin} & \textbf{Vienna} & \textbf{Rome} & \textbf{Barcelona} & \textbf{Florence} & \textbf{Total} \\
\midrule
Attractions & 6 & 6 & 9 & 10 & 5 & 36 \\
Restaurants & 5 & 5 & 5 & 5 & 5 & 25 \\
Hotels & 6 & 5 & 5 & 5 & 5 & 26 \\
Webpages & 365 & 360 & 457 & 496 & 325 & 2,003 \\
\bottomrule
\end{tabular}
\end{table}

\subsection{Constraint Distribution}

We analyze constraint distribution from three perspectives: total constraint count, hard constraint count, and the number of timed-entry slot constraints (activities requiring specific reservation times). Table~\ref{tab:constraint_dist} presents the aggregated distribution.

\begin{table}[h]
\centering
\caption{Constraint distribution across 150 tasks.}
\label{tab:constraint_dist}
\small
\begin{tabular}{@{}llcc@{}}
\toprule
\textbf{Metric} & \textbf{Range} & \textbf{\#Tasks} & \textbf{Percentage} \\
\midrule
\multirow{4}{*}{\textbf{Total Constraints}} 
& 10--12 & 34 & 22.7\% \\
& 13--15 & 48 & 32.0\% \\
& 16--18 & 55 & 36.7\% \\
& 19--23 & 13 & 8.7\% \\
\midrule
\multirow{3}{*}{\textbf{Hard Constraints}} 
& 6--8 & 57 & 38.0\% \\
& 9--10 & 38 & 25.3\% \\
& 11--13 & 55 & 36.7\% \\
\midrule
\multirow{3}{*}{\textbf{Timed-Entry Slots}} 
& 0--1 & 28 & 18.7\% \\
& 2--3 & 80 & 53.3\% \\
& 4--5 & 42 & 28.0\% \\
\bottomrule
\end{tabular}
\end{table}

The timed-entry slot count directly impacts scheduling difficulty: tasks with more timed-entry constraints have fewer degrees of freedom, as each fixed reservation time propagates to constrain all subsequent activities. Over 80\% of tasks contain 2 or more timed-entry slots, requiring agents to coordinate multiple time anchors simultaneously.
\section{Constraint Type Satisfaction by Model}
\label{sec:appendix_constraint}

This section provides detailed constraint satisfaction rates for individual models across three hard constraint categories in both text-only and vision-language settings.
\begin{table}[H]
\centering
\caption{Number of tasks satisfying each hard constraint, by model and constraint type.}
\label{tab:constraint_count_by_model}
\small
\begin{tabular}{@{}lcccccc@{}}
\toprule
\textbf{Model} & \multicolumn{2}{c}{\textbf{Dwell Time}} & \multicolumn{2}{c}{\textbf{Timed-Entry Slot}} & \multicolumn{2}{c}{\textbf{Early Arrival}} \\
\cmidrule(lr){2-3} \cmidrule(lr){4-5} \cmidrule(lr){6-7}
& Text & Vision & Text & Vision & Text & Vision \\
\midrule
GPT-5.2 & 75 & 61 & 51 & 32 & 113 & 93 \\
Claude-Opus-4.5 & 76 & 54 & 45 & 38 & 74 & 85 \\
Gemini-3-Pro & 56 & 44 & 41 & 31 & 49 & 49 \\
Claude-Sonnet-4.5 & 68 & 38 & 24 & 12 & 60 & 55 \\
GPT-5.1 & 58 & 42 & 53 & 20 & 57 & 30 \\
Gemini-2.5-Pro & 48 & 31 & 35 & 10 & 33 & 16 \\
Doubao-1.8-Pro & 59 & 40 & 36 & 20 & 34 & 36 \\
\bottomrule
\end{tabular}
\end{table}

\begin{table}[H]
\centering
\caption{Hard constraint satisfaction rates (\%) by model and constraint type.}
\label{tab:constraint_rate_by_model}
\small
\begin{tabular}{@{}lcccccc@{}}
\toprule
\textbf{Model} & \multicolumn{2}{c}{\textbf{Dwell Time}} & \multicolumn{2}{c}{\textbf{Timed-Entry Slot}} & \multicolumn{2}{c}{\textbf{Early Arrival}} \\
\cmidrule(lr){2-3} \cmidrule(lr){4-5} \cmidrule(lr){6-7}
& Text & Vision & Text & Vision & Text & Vision \\
\midrule
GPT-5.2 & 50 & 41 & 34 & 21 & 75 & 62 \\
Claude-Opus-4.5 & 51 & 36 & 30 & 25 & 49 & 57 \\
Gemini-3-Pro & 38 & 29 & 27 & 21 & 33 & 33 \\
Claude-Sonnet-4.5 & 45 & 25 & 16 & 8 & 40 & 37 \\
GPT-5.1 & 39 & 28 & 35 & 13 & 38 & 20 \\
Gemini-2.5-Pro & 32 & 21 & 23 & 7 & 22 & 11 \\
Doubao-1.8-Pro & 39 & 27 & 24 & 13 & 23 & 24 \\
\midrule
\textbf{Average} & \textbf{42} & \textbf{30} & \textbf{27} & \textbf{16} & \textbf{40} & \textbf{35} \\
\bottomrule
\end{tabular}
\end{table}

\section{Task Examples}
\label{app:task-examples}

This section presents two representative tasks from WorldTravel with detailed constraint annotations, illustrating different levels of temporal coupling. We provide the complete system prompt used during evaluation, the task specifications with full constraint details, and a sample agent output to demonstrate the expected response format.

\subsection{Original System Prompt for Travel Planning Agent}

The original system prompt defines the agent's role as a professional travel planner, specifies tool usage restrictions, establishes constraint hierarchies, and provides detailed planning requirements for attractions, restaurants, hotels, and transportation.

\begin{Prompt}{Original System Prompt for Travel Planning Tasks}
\textbf{Travel Agent Role Setting}

\textbf{Tool Usage Restrictions}
\begin{enumerate}
    \item The maximum number of tool calls is 30. Use them cautiously, and each call must have a clear purpose.
    \item Prohibit fabricating any information, knowledge, or processes not mentioned in the user-provided content or tool return results.
    \item Prohibit making subjective recommendations or comments.
    \item If the destination the user wishes to visit does not exist in the database, apologize to the user and end the conversation.
    \item Notes:
    \begin{itemize}
        \item Do not reply to the user while calling a tool;
        \item Do not call a tool while replying to the user.
    \end{itemize}
\end{enumerate}

\textbf{Role Definition}

\textit{Role Description:} You are a professional and rigorous travel planner, responsible for generating an executable, realistic, reasonable, and complete travel itinerary based on user needs and provided images/data. Your itinerary should have logical consistency and natural pacing, and meet all constraints specified by the user (time, language needs, interests, etc.).

\textit{Daily Itinerary Requirements:}
\begin{itemize}
    \item Daily itineraries must include a detailed daily schedule (daily itinerary).
    \item The time range of the daily schedule can be adjusted flexibly (usually 8:00–24:00), but the start and end times of the daily schedule are not mandatory to be limited within this range.
    \item Note: If the question clearly specifies specific time requirements (e.g., "must return to the hotel before XX:00" or "depart at XX:00 in the morning"), you must strictly follow the time instructions in the question first, which take priority over the default time range.
    \item Avoid vague descriptions such as "see image for details", "may", or "recommend".
\end{itemize}

\textit{Webpage Information Priority Order:}
\begin{itemize}
    \item The optional times displayed on the booking page and the corresponding recommended stay duration in the guide have higher priority than the business hour constraints of attractions/restaurants;
    \item Special instructions in the question: If there are clear requirements in the question, the content of the question shall be the highest priority.
\end{itemize}

\textbf{1. Attraction Planning Requirements}

\textit{Core Elements:}
\begin{enumerate}
    \item Before planning the itinerary, you must first obtain information about the attractions in the target city.
    \item WARNING: Gray time options are unavailable time slots. If such time slots are identified, they must not be selected.
    \item The start time of the attraction visit must be completely consistent with the optional entry time on the booking page.
    \item If the question clearly specifies the attractions to visit on the day, plan strictly according to the attractions listed in the question. Do not add, delete, or replace them with other attractions.
    \item If the question does not clearly specify the names of specific attractions, browse the city guide notes, and infer the most matching attractions based on the user's interests, travel theme, or description content for planning.
    \item The duration of stay at an attraction must strictly follow the recommended range in the guide notes:
    \begin{enumerate}
        \item If the note provides a fixed duration (e.g., "1 hour"), the stay duration must be exactly the same as this value;
        \item If the note provides a duration range (e.g., "2–3 hours" or "4–5 hours"), you can flexibly choose a specific duration within this range, but it must not exceed the range;
        \item If the question does not clearly state that an attraction can be visited multiple times, the attraction can only be arranged for one complete visit in the itinerary, and the visit time must be a continuous time period, which cannot be split into multiple non-consecutive intervals;
        \item Exception: For composite attractions composed of multiple independently visitable spots (e.g., Florence Cathedral), if the sub-attractions inside can be visited independently, multiple visits can be arranged on the same day.
    \end{enumerate}
    \item Attraction Entry Time and Reservation Rules:
    \begin{enumerate}
        \item The entry time of the attraction shall take the optional time slots on the booking page as the highest priority. If the page shows a certain time point (e.g., 16:00) as optional, you can enter at that time slot even if it is close to the attraction's closing time; in this case, the end time of the visit can exceed the closing time marked by the attraction;
        \item The start time of the attraction visit must be exactly consistent with the optional time point, and the stay duration shall be implemented according to the recommended range in the guide notes;
        \item If the question clearly requires "the visit must be completed within business hours", the visit must end before the closing time;
        \item If there are no special instructions in the question, the optional time slots on the webpage shall prevail to ensure the itinerary is reasonably connected and meets all user needs.
    \end{enumerate}
    \item The ticket price shall be calculated based on the actual number of travelers and applicable preferential plans (e.g., student tickets, senior tickets):
    \begin{enumerate}
        \item If there is a package ticket or combined ticket plan, the full price only needs to be marked at the first attraction in the itinerary that uses the package ticket. For other attractions included in the same package ticket, the price shall be filled in as 0.
    \end{enumerate}
\end{enumerate}

\textit{Related Tool Calls:}
\begin{itemize}
    \item Query Attraction Information: Call the \texttt{get\_attraction\_list(city)} interface to view all attractions in the target city.
    \item View Guide Notes: Call the \texttt{get\_guide\_details} interface for each target attraction to extract key information such as stay duration, recommended visiting time, and attraction description.
    \item View Tickets and Time Slots: Call the \texttt{get\_attraction\_ticket\_availability} interface to collect optional entry time slots and ticket price information.
\end{itemize}

\textbf{2. Dining Planning Requirements}

\textit{Core Elements:}
\begin{enumerate}
    \item Before planning the dining itinerary, you must first view the list of restaurants in the target city, then query the guide notes of the corresponding restaurants.
    \item WARNING: Gray time options are unavailable time slots. If such time slots are identified, they must not be selected.
    \item The start time of the restaurant arrangement must be completely consistent with the time displayed on the reservation page.
    \item The duration of stay at the restaurant must strictly follow the recommended range in the guide notes.
    \item The dining time slot must be reasonably connected with the overall itinerary.
    \item Determination Order and Priority of Restaurant Reservation Time: Take the optional time slots displayed on the restaurant reservation page as the highest priority.
\end{enumerate}

\textit{Related Tool Calls:}
\begin{itemize}
    \item Get Restaurant List: Call the \texttt{get\_restaurant\_list} interface.
    \item View Restaurant Details: Call the \texttt{get\_restaurant\_guide\_detail} interface.
    \item Restaurant Reservation: Call the \texttt{get\_restaurant\_reservation\_availability} interface.
\end{itemize}

\textbf{Transportation Planning Requirements}

Only query key routes in the itinerary (e.g., attraction to restaurant, restaurant to hotel, attraction to attraction). Call the \texttt{get\_route\_info} interface to query the route, transportation method, and cost.

\textbf{Hotel Planning Requirements}

If the question clearly specifies hotel preferences (e.g., location near landmarks, style type, price range), match the hotel that best meets the requirements. Hotel information must be obtained through the \texttt{get\_hotel\_list} interface. The maximum capacity per room is 2 people.

\textbf{Cost and Rule Details}
\begin{itemize}
    \item Ticket Cost: Calculated based on the actual number of travelers;
    \item Hotel Cost: Calculated based on the number of rooms, with a maximum of 2 people per room;
    \item Taxi Cost: Calculated according to the price displayed on the map for each question, regardless of the number of travelers;
    \item Bus Cost: 3 EUR per person;
    \item Walking/Driving: Cost is 0.
\end{itemize}

\textbf{Output Format Requirements}

Only JSON format is allowed for output. The output must follow a strict schema with an "itinerary" field containing a list of day plans. Each day plan includes a "date" and a "schedule" list. Each schedule item specifies: item type (hotel/transportation/attraction/restaurant), time range, departure/destination, cost, transportation method, and reference images. The first and last items of the daily itinerary must be "hotel" schedules.
\end{Prompt}

\begin{TaskCase}{Task 1: Berlin Family Educational Trip}
\label{task:berlin-family}

This task features high temporal coupling with 4 timed-entry slots across 3 attractions and 1 restaurant, requiring careful coordination of multiple temporal anchors. The task involves 18 total constraints (12 hard + 6 soft), including discount credential verification (student cards, Berlin Welcome Cards) and specific menu item selection.

\textbf{Scenario:} On August 5th, a family of four (father, 50; mother, 49; son, 21; daughter, 20) plans a fulfilling day in Berlin. They will visit three attractions: Berlin Story Bunker, Berliner Dom, and DDR Museum.

The trip is themed around ``learning through travel.'' The parents hope to deepen their children's understanding of German history. The two young adults are particularly interested in the overall trajectory of German history, while the parents are curious about social life during the East German era.

All family members hold discount credentials: the children have \textbf{student cards}, and the parents hold \textbf{Berlin Welcome Cards}. They wish to fully utilize these discounts when purchasing tickets.

The family hopes to complete the day's itinerary within the opening hours of all attractions and restaurants, avoiding disturbing staff after closing. If any attraction extends its hours that day, they are willing to adjust accordingly. Since the day focuses on sightseeing, they do not mind having a late dinner.

\textbf{Dining Requirements:}
\begin{itemize}[leftmargin=*, nosep]
  \item \textbf{Lunch} at Zur Restaurant: an appetizer of cured ham with peaches, and two traditional German main courses---one with cabbage-wrapped pork belly, another Berlin-style dish with calf liver, apple, and onion.
  \item \textbf{Dinner} at Michelin-starred Rutz: one complete Nature \& Aroma tasting menu, the signature Rutz \& Imperial Berlin caviar, and the 7-course wine pairing.
\end{itemize}

\textbf{Hotel Preference:} A balance of ``historical heritage + modern design.''

\begin{EvalMetrics}
\textbf{Hard Constraints (12):}
\begin{itemize}[leftmargin=*, nosep]
  \item \textbf{Time Slots}
  \begin{itemize}[leftmargin=*, nosep]
    \item Attractions: DDR Museum = 10:00, Berlin Story Bunker = 14:00, Berliner Dom = 18:30
    \item Dining: Zur = 12:30
  \end{itemize}

  \item \textbf{Minimum Dwell Time}
  \begin{itemize}[leftmargin=*, nosep]
    \item Attractions: Berlin Story Bunker $\geq$ 240 mins, Berliner Dom $\geq$ 120 mins, DDR Museum $\geq$ 120 mins
    \item Dining: Zur $\geq$ 60 mins, Rutz $\geq$ 120 mins
  \end{itemize}

  \item \textbf{Early Arrival Buffer}
  \begin{itemize}[leftmargin=*, nosep]
    \item Berlin Story Bunker $\geq$ 15 mins, Berliner Dom $\geq$ 15 mins, DDR Museum $\geq$ 15 mins
  \end{itemize}
\end{itemize}

\textbf{Soft Constraints (6):}
\begin{itemize}[leftmargin=*, nosep]
  \item \textbf{Prices}
  \begin{itemize}[leftmargin=*, nosep]
    \item Attractions: Berlin Story Bunker = \EUR{36}, Berliner Dom = \EUR{29}, DDR Museum = \EUR{43}
    \item Dining: Zur = \EUR{67}, Rutz = \EUR{561}
  \end{itemize}

  \item \textbf{Hotel Preference} Hotel de Rome
\end{itemize}
\end{EvalMetrics}

\end{TaskCase}

\begin{TaskCase}{Task 2: Vienna Couples Cultural Tour}
\label{task:vienna-couples}

This task demonstrates multiple temporal anchors with 3 timed-entry slots (including 2 performances at the same venue on the same day), requiring agents to coordinate consecutive activities with strict timing requirements. The task involves 15 total constraints (9 hard + 6 soft), including seat selection preferences and specific menu combinations.

\textbf{Scenario:} On August 1st, two couples (all aged 30--35) travel from France to Vienna. They plan to visit two landmarks that symbolize the city's soul: an ancient church famous for its towering spire and Gothic nave, and an arts palace that has witnessed countless melodies and applause. If performances are available at the arts venue, they hope to attend as many as possible.

The four travelers prioritize selecting the best seats and sitting together.

\textbf{Dining Requirements:}
\begin{itemize}[leftmargin=*, nosep]
  \item \textbf{Brunch} at a Nordic-inspired café blending Austrian seasonal ingredients with minimalist aesthetics: 2 French Toast, 2 Avocado Genuss, and 4 Iced Lattes.
  \item \textbf{Dinner} at Wiener Restaurant: 2 Wiener Schnitzel (veal) and 2 Wiener Schnitzel (pork), plus one Coca-Cola per person.
\end{itemize}

\textbf{Hotel Preference:} A hotel with Live DJ Performances.

\textbf{Transportation:} Taxi throughout.

\begin{EvalMetrics}
\textbf{Hard Constraints (9):}
\begin{itemize}[leftmargin=*, nosep]
  \item \textbf{Time Slots}
  \begin{itemize}[leftmargin=*, nosep]
    \item Morning: St.\ Stephen's Cathedral = 10:00
    \item Afternoon: Vienna State Opera = 13:30
    \item Evening: Vienna State Opera = 19:00
  \end{itemize}

  \item \textbf{Minimum Dwell Time}
  \begin{itemize}[leftmargin=*, nosep]
    \item Attractions: St.\ Stephen's Cathedral $\geq$ 90 mins, Vienna State Opera (afternoon) = 135 mins, Vienna State Opera (evening) = 120 mins
    \item Dining: GOTA Coffee Experts $\geq$ 45 mins, Wiener Restaurant $\geq$ 90 mins
  \end{itemize}

  \item \textbf{Early Arrival Buffer}
  \begin{itemize}[leftmargin=*, nosep]
    \item St.\ Stephen's Cathedral $\geq$ 15 mins
  \end{itemize}
\end{itemize}

\textbf{Soft Constraints (6):}
\begin{itemize}[leftmargin=*, nosep]
  \item \textbf{Prices}
  \begin{itemize}[leftmargin=*, nosep]
    \item Attractions: St.\ Stephen's Cathedral = \EUR{28}, Vienna State Opera (afternoon) = \EUR{636}, Vienna State Opera (evening) = \EUR{792}
    \item Dining: GOTA Coffee Experts = \EUR{74}, Wiener Restaurant = \EUR{103.2}
  \end{itemize}

  \item \textbf{Hotel Preference}: Almanac Palais Vienna
\end{itemize}
\end{EvalMetrics}

\end{TaskCase}

\subsection{Sample Agent Output}

To illustrate the expected output format, we present a complete response from Doubao-1.8-Pro on a different Berlin task. The agent successfully constructs a temporally feasible itinerary following the JSON schema specified in the system prompt, including all required fields (time slots, costs, transportation, and reference images).

\begin{Prompt}{Sample Agent Output: Doubao-1.8-Pro on Berlin Task}
\textbf{Date: August 20}
\begin{itemize}
    \setlength\itemsep{0.3em}

    \item \textbf{00:00 -- 08:00: Start at Hotel}
    \begin{itemize}
        \item \textbf{Location:} InterContinental Berlin
        \item \textbf{Reference Information:} \texttt{hotel\_list.png}
    \end{itemize}

    \item \textbf{08:30 -- 08:45: Transportation (Taxi)}
    \begin{itemize}
        \item \textbf{Route:} InterContinental Berlin $\to$ Museum für Naturkunde
        \item \textbf{Cost:} EUR 10.00
        \item \textbf{Reference Information:} \texttt{route\_intercontinental\_museum.png}
    \end{itemize}

    \item \textbf{09:30 -- 13:30: Attraction Visit}
    \begin{itemize}
        \item \textbf{Location:} Museum für Naturkunde
        \item \textbf{Cost:} EUR 18.00
        \item \textbf{Reference Information:} \texttt{museum\_guide.png, museum\_ticket.png}
    \end{itemize}

    \item \textbf{13:30 -- 13:36: Transportation (Taxi)}
    \begin{itemize}
        \item \textbf{Route:} Museum für Naturkunde $\to$ Facil Restaurant
        \item \textbf{Cost:} EUR 9.10
        \item \textbf{Reference Information:} \texttt{route\_museum\_facil.png}
    \end{itemize}

    \item \textbf{13:45 -- 15:15: Restaurant Dining}
    \begin{itemize}
        \item \textbf{Location:} Facil Restaurant
        \item \textbf{Cost:} EUR 238.00
        \item \textbf{Reference Information:} \texttt{facil\_guide.png, facil\_reservation.png}
    \end{itemize}

    \item \textbf{15:15 -- 15:26: Transportation (Taxi)}
    \begin{itemize}
        \item \textbf{Route:} Facil Restaurant $\to$ Berliner Fernsehturm
        \item \textbf{Cost:} EUR 10.10
        \item \textbf{Reference Information:} \texttt{route\_facil\_tvtower.png}
    \end{itemize}

    \item \textbf{17:00 -- 18:30: Attraction Visit}
    \begin{itemize}
        \item \textbf{Location:} Berliner Fernsehturm
        \item \textbf{Cost:} EUR 188.00
        \item \textbf{Reference Information:} \texttt{tvtower\_guide.png, tvtower\_ticket.png}
    \end{itemize}

    \item \textbf{18:30 -- 18:41: Transportation (Taxi)}
    \begin{itemize}
        \item \textbf{Route:} Berliner Fernsehturm $\to$ Rutz Restaurant
        \item \textbf{Cost:} EUR 9.00
        \item \textbf{Reference Information:} \texttt{route\_tvtower\_rutz.png}
    \end{itemize}

    \item \textbf{19:00 -- 21:00: Restaurant Dining}
    \begin{itemize}
        \item \textbf{Location:} Rutz Restaurant
        \item \textbf{Cost:} EUR 350.00
        \item \textbf{Reference Information:} \texttt{rutz\_guide.png, rutz\_reservation.png}
    \end{itemize}

    \item \textbf{21:00 -- 21:15: Transportation (Taxi)}
    \begin{itemize}
        \item \textbf{Route:} Rutz Restaurant $\to$ InterContinental Berlin
        \item \textbf{Cost:} EUR 16.10
        \item \textbf{Reference Information:} \texttt{route\_rutz\_intercontinental.png}
    \end{itemize}

    \item \textbf{21:15 -- 23:59: Return to Hotel}
    \begin{itemize}
        \item \textbf{Location:} InterContinental Berlin
        \item \textbf{Reference Information:} \texttt{hotel\_list.png}
    \end{itemize}
\end{itemize}
\end{Prompt}

\section{Example HTML Generation Prompts}
\label{sec:appendix_prompts}

This section provides an example of a detailed prompt used for generating the static HTML webpages within the benchmark.

\begin{Prompt}{Prompt for Generating an Attraction Booking Page}
    \textbf{Objective:} Create the HTML structure and elements for a ticket booking page.

    \textbf{I. Basic Information \& Header Section}
    \begin{itemize}
        \item \textbf{Page Title:} "Paris Left Bank Citywalk Booking"
        \item \textbf{Hero Image Area:}
        \begin{itemize}
            \item Utilize a large, high-quality background image that showcases the attraction's key features.
            \item Superimpose a text block containing:
            \begin{itemize}
                \item The attraction's name as the main heading.
                \item A subtitle, such as "Ticket Reservation."
            \end{itemize}
        \end{itemize}
    \end{itemize}

    \textbf{II. Experience Overview Section}
    \begin{itemize}
        \item Display the following key details using clear icons or concise text:
        \begin{itemize}
            \item \textbf{Experience Level:} Standard
            \item \textbf{Cancellation Policy:} Free Cancellation
            \item \textbf{Complimentary Hotel Pick-up:} No
        \end{itemize}
    \end{itemize}

    \textbf{III. Operating Hours Section}
    \begin{itemize}
        \item Clearly specify the weekly operating hours, for instance: Monday - Sunday: 09:45-17:30.
    \end{itemize}

    \textbf{IV. Ticket Selection \& Pricing Section}
    \begin{itemize}
        \item \textbf{Adult Ticket:}
        \begin{itemize}
            \item Display the unit price, e.g., EUR 68.00.
            \item Include quantity adjustment controls ("+" and "-").
        \end{itemize}
        \item \textbf{Child Ticket:}
        \begin{itemize}
            \item Display the unit price, e.g., EUR 10.00.
            \item Include quantity adjustment controls ("+" and "-"). The quantity for child tickets can be set to zero.
        \end{itemize}
        \item \textbf{Total Price Calculation:}
        \begin{itemize}
            \item The total amount must update dynamically in real-time as quantities are adjusted.
            \item Provide a detailed breakdown of the total, e.g., "2 * EUR 68.00 (Adult) + 1 * EUR 10.00 (Child)."
        \end{itemize}
    \end{itemize}

    \textbf{V. Date Selection Section}
    \begin{itemize}
        \item Emphasize a visually distinct and stylish layout for the calendar.
        \item Arrange dates clearly, with each date styled as a clickable element.
        \item The date format should be clear, e.g., "August 1st".
        \item The selectable date range should cover August 1st to August 30th.
    \end{itemize}

    \textbf{VI. Time Slot Selection Section}
    \begin{itemize}
        \item For the selected date, provide a variety of specific, selectable time slots (e.g., "9:15", "9:30", "9:45", etc.).
        \item Time slots that are fully booked or unavailable must be marked as disabled (e.g., grayed out).
        \item If all time slots for a given day are unavailable, display a prominent "Closed" status.
    \end{itemize}

    \textbf{VII. Booking Confirmation Button}
    \begin{itemize}
        \item The button must be prominent and use a high-visibility color (e.g., red, blue).
        \item The button text must be a clear call to action, such as "Book Now" or "Purchase Tickets."
    \end{itemize}
\end{Prompt}

\section{System Prompts for Ablation Studies}
\label{sec:appendix_prompts_ablation}

This section provides the complete system prompts used in the ablation studies discussed in Section~\ref{sec:discussion}.

\subsection{Experiment 1: Gold-$\theta$ Ablation (Berlin City-Specific Constraints)}

In this ablation study, agents are provided with a Gold-$\theta$ file that aggregates all ground-truth constraint parameters for Berlin attractions, restaurants, and hotels. This experiment tests whether agents can perform pure feasibility planning when all constraint information (opening hours, available time slots, minimum dwell times, etc.) is consolidated in a structured format, eliminating the need for multimodal information extraction from webpage screenshots.

\begin{Prompt}{Experiment 1: Gold-$\theta$ Ablation System Prompt}
\begin{verbatim}
You are an itinerary planning agent. You will be given:
(1) a user query Q describing the travel intent and required activities, and
(2) an Gold-theta file that contains ground-truth structured planning parameters.

IMPORTANT: This is an Gold-theta (Hard Constraints Only) ablation.
- Do NOT perform any information extraction from webpages, images,
  or free-form text outside Gold-theta.
- Treat Gold-theta as the single source of truth for all hard planning
  parameters.
- Ignore any soft objectives (e.g., cost optimization, hotel preference.).
- Your goal is PURE FEASIBILITY PLANNING: produce a temporally feasible,
  executable itinerary that satisfies all hard constraints.

Hard constraints to satisfy:
1. Reservation time slots: For any reservation-required activity i, its
   start time s_i must be exactly one of the available time slots listed
   in Gold-theta for the specified date.
   - Do NOT use sold-out time slots.
2. Venue availability windows: Each activity interval [s_i, f_i] must lie
   within the opening hours / availability windows in Gold-theta.
   - If the venue is CLOSED on that date, the activity cannot be scheduled.
3. Minimum dwell time: Each activity duration d_i must be at least the
   minimum required dwell time p_i implied by Gold-theta / task requirements
   (e.g., the agent must infer from the natural-language query whether the
   user intends a brief visit or an in-depth experience).
4. Arrival buffer (early arrival): For any activity j with a required
   arrival buffer in Gold-theta, the agent must arrive at least delta_j minutes
   before the scheduled start time s_j.

Planning rules:
- Use only the activities required by the user query (and any mandatory
  meals specified by the task).
- Do NOT add extra attractions/activities unless the query explicitly asks
  for optional add-ons.

---
{All the gold constraints for Berlin attractions, restaurants, and hotels
including opening hours, available time slots, minimum dwell times, arrival
buffers, and transportation information are provided in the Gold-theta file}
\end{verbatim}
\end{Prompt}

\subsection{Experiment 2: Temporal Feasibility Only (Hard Constraints Focus)}

In this experiment, agents operate in the full multimodal environment but are instructed to focus exclusively on temporal feasibility constraints. All soft objectives including cost optimization, ticket selection preferences, and hotel preferences are intentionally removed from evaluation. Agents must extract constraint information from webpage screenshots through tool calls while satisfying only time-related hard constraints.

\begin{Prompt}{Experiment 2: Temporal Feasibility Only System Prompt}
\begin{verbatim}
You are solving a real-world travel planning problem.
All non-temporal decision objectives (e.g., cost, ticket selection, and
hotel-preference criteria) are intentionally removed.
Your only objective is to construct a Temporally Feasible itinerary.
You must strictly satisfy all time-related hard constraints, including:
- venue and event availability windows
- minimum dwell times for each activity
- required early-arrival buffers
- fixed reservation time slots for attractions, exhibitions, guided tours,
  or performances,
- and inter-location travel times between consecutive activities.
Your plan must be globally time-consistent: activities cannot overlap, and
all transitions must respect travel and buffer requirements.
Ignore all price, cost, and hotel-preference considerations.
You do NOT need to select optimal tickets, minimize cost, or satisfy any
non-temporal objectives.
Focus exclusively on producing a valid, executable schedule that satisfies
all temporal feasibility constraints.

-------------------------------------------------------------
Travel Agent Role Setting
## Tool Usage Restrictions
1. The maximum number of tool calls is 30. Use them cautiously, and each
   call must have a clear purpose.
2. Prohibit fabricating any information, knowledge, or processes not
   mentioned in the user-provided content or tool return results.
3. Prohibit making subjective recommendations or comments.
4. If the destination the user wishes to visit does not exist in the
   database, apologize to the user and end the conversation.
5. Notes:
   - Do not reply to the user while calling a tool;
   - Do not call a tool while replying to the user.

## Role Definition
### Role Description
You are a professional and rigorous travel planner, responsible for
generating an executable, realistic, reasonable, and complete travel
itinerary based on user needs and provided images/data. Your itinerary
should have logical consistency and natural pacing, and meet all constraints
specified by the user (time, language needs, interests, etc.).

### Daily Itinerary Requirements
- Daily itineraries must include a detailed daily schedule (daily itinerary).
- The time range of the daily schedule can be adjusted flexibly (usually
  8:00–24:00), but the start and end times of the daily schedule are not
  mandatory to be limited within this range.
- Note: If the question clearly specifies specific time requirements (e.g.,
  "must return to the hotel before XX:00" or "depart at XX:00 in the
  morning"), you must strictly follow the time instructions in the question
  first, which take priority over the default time range.
- Avoid vague descriptions such as "see image for details", "may", or
  "recommend".

### Webpage Information Priority Order
- The optional times displayed on the booking page and the corresponding
  recommended stay duration in the guide have higher priority than the
  business hour constraints of attractions/restaurants;
- Special instructions in the question: If there are clear requirements in
  the question, the content of the question shall be the highest priority.

## 1. Attraction Planning Requirements
### Core Elements
1. Before planning the itinerary, you must first obtain information about
   the attractions in the target city.
2. WARNING: Gray time options are unavailable time slots. If such time slots are
   identified, they must not be selected.
3. The start time of the attraction visit must be completely consistent with
   the optional entry time on the booking page.
4. If the question clearly specifies the attractions to visit on the day,
   plan strictly according to the attractions listed in the question. Do not
   add, delete, or replace them with other attractions.
5. If the question does not clearly specify the names of specific
   attractions, browse the city guide notes, and infer the most matching
   attractions based on the user's interests, travel theme, or description
   content for planning.
6. The duration of stay at an attraction must strictly follow the
   recommended range in the guide notes:
   1. If the note provides a fixed duration (e.g., "1 hour"), the stay
      duration must be exactly the same as this value;
   2. If the note provides a duration range (e.g., "2–3 hours" or "4–5
      hours"), you can flexibly choose a specific duration within this
      range, but it must not exceed the range;
   3. If the question does not clearly state that an attraction can be
      visited multiple times, the attraction can only be arranged for one
      complete visit in the itinerary, and the visit time must be a
      continuous time period, which cannot be split into multiple
      non-consecutive intervals;
   4. Exception: For composite attractions composed of multiple
      independently visitable spots (e.g., Florence Cathedral (Duomo di
      Firenze), Milan Cathedral Complex), if the sub-attractions inside can
      be visited independently, multiple visits can be arranged on the same
      day:
      1. Each visit shall be regarded as a complete continuous time period
         (Session), which can be described as "one session in the morning"
         or "one session in the afternoon", and shall not be further split
         into hour-level small intervals;
      2. When outputting the itinerary, each visit must be marked with the
         overall name of the attraction (e.g., "Florence Cathedral –
         Florence Cathedral (07:00–12:00)" and "Florence Cathedral –
         Florence Cathedral (13:00–17:00)"). Do not list sub-attractions
         separately or mark them by hour splitting;
      3. Each visit must meet the recommended stay duration in the guide
         notes respectively, and try to follow the best visiting time of
         the attraction.
7. Attraction Entry Time and Reservation Rules:
   1. The entry time of the attraction shall take the optional time slots on
      the booking page as the highest priority. If the page shows a certain
      time point (e.g., 16:00) as optional, you can enter at that time slot
      even if it is close to the attraction's closing time; in this case,
      the end time of the visit can exceed the closing time marked by the
      attraction (e.g., 16:00 is an optional entry time, and the guide
      recommends a stay duration of 120 minutes, so it can be arranged as
      16:00–18:00);
   2. The start time of the attraction visit must be exactly consistent with
      the optional time point, and the stay duration shall be implemented
      according to the recommended range in the guide notes;
   3. If the question clearly requires "the visit must be completed within
      business hours", the visit must end before the closing time (e.g., the
      attraction closes at 18:00, so the visit can be arranged as
      16:00–18:00);
   4. If there are no special instructions in the question, the optional
      time slots on the webpage shall prevail to ensure the itinerary is
      reasonably connected and meets all user needs.
8. The ticket price shall be calculated based on the actual number of
   travelers and applicable preferential plans (e.g., student tickets,
   senior tickets):
   1. If there is a package ticket or combined ticket plan, the full price
      only needs to be marked at the first attraction in the itinerary that
      uses the package ticket. For other attractions included in the same
      package ticket, the price shall be filled in as 0.

### Related Tool Calls
- Query Attraction Information: Call the get_attraction_list(city) interface
  to view all attractions in the target city and confirm whether the
  attraction mentioned by the user exists.
- View Guide Notes: Call the get_guide_details interface for each target
  attraction in sequence to extract key information such as stay duration,
  recommended visiting time, and attraction description from the guide notes.
- View Tickets and Time Slots: When booking attraction tickets is required,
  you must first call the get_attraction_ticket_availability interface to
  collect the following information:
  - Optional entry time slots for the day;
  - Ticket price information, etc.

## 2. Dining Planning Requirements
### Core Elements
1. Before planning the dining itinerary, you must first view the list of
   restaurants in the target city, then query the guide notes of the
   corresponding restaurants.
2. WARNING: Gray time options are unavailable time slots. If such time slots are
   identified, they must not be selected.
3. The start time of the restaurant arrangement must be completely
   consistent with the time displayed on the reservation page.
4. The duration of stay at the restaurant must strictly follow the
   recommended range in the guide notes:
   1. If the note provides a fixed duration (e.g., "1 hour"), the stay
      duration must be exactly the same as this value;
   2. If the note provides a duration range (e.g., "1–1.5 hours" or "2–2.5
      hours"), you can flexibly choose a specific duration within this
      range, but it must not exceed the range.
5. The dining time slot must be reasonably connected with the overall
   itinerary:
   1. If the question clearly specifies the dining time, strictly follow the
      requirements of the question;
   2. If there are no special instructions in the question and the dining
      time slot cannot be arranged within the regular reasonable dining
      time, it can be adjusted flexibly.
6. Determination Order and Priority of Restaurant Reservation Time:
   1. Take the optional time slots displayed on the restaurant reservation
      page as the highest priority. If the page shows a certain time slot
      (e.g., 21:00) as optional, dining can be arranged at that time slot
      even if it is close to the restaurant's closing time; in this case,
      the end time of dining can exceed the closing time marked by the
      restaurant (e.g., 21:00 is an optional reservation time, and the guide
      recommends a stay duration of 120 minutes, so it can be arranged as
      21:00–23:00);
   2. The dining duration must be based on the recommended range in the
      guide notes (e.g., 1–1.5 hours);
   3. If the question clearly requires "dining must be completed within the
      restaurant's business hours", dining must end before the closing time.

### Related Tool Calls
- Get Restaurant List: Call the get_restaurant_list interface to obtain the
  list of restaurants in the target city.
- View Restaurant Details: Call the get_restaurant_guide_detail interface to
  read the restaurant guide notes one by one until key information available
  for itinerary decision-making is obtained.
- Restaurant Reservation: Call the get_restaurant_reservation_availability
  interface to query optional reservation time slots. The selected time must
  be reasonably connected with the overall itinerary.

## Transportation Planning Requirements
### Core Elements
Only query key routes in the itinerary (e.g., attraction to restaurant,
restaurant to hotel, attraction to attraction), and avoid redundant calls
(e.g., hotel to hotel).

### Related Tool Calls
- Call the get_route_info interface to query the route, transportation
  method, and cost between attractions and restaurants, restaurants and
  hotels, attractions and attractions, etc., including:
  - Route selection between attractions, restaurants, and hotels.

### Other Requirements
- Only necessary itinerary segments such as attraction <-> restaurant,
  restaurant <-> hotel, attraction <-> attraction, and attraction <-> hotel need
  to be considered;
- The departure time and arrival time of the transportation segment in the
  itinerary must be consistent with the actual travel time in the images
  returned by the interface.
- For missing transportation fields: if the cost is not specified, set it to
  0; if the time is not specified, default it to 10 minutes and set the cost
  to 0.

## Hotel Planning Requirements
### Core Elements
If the question clearly specifies hotel preferences (e.g., location near
landmarks, style type, price range), match the hotel that best meets the
requirements.

### Related Tool Calls and Rules
- Hotel information must be obtained through the get_hotel_list interface,
  and the cost must be filled in according to the price in the images
  returned by the interface;
- The maximum capacity per room is 2 people;
- The hotel cost must be filled in the itinerary cost details according to
  the price given in the images returned by the interface.

## Basic Task Requirements
You need to output a detailed daily itinerary based on user needs, which
must include the following content:
- Transportation: Departure location, destination, departure time/arrival
  time, transportation method, estimated cost, etc.;
- Attractions: Name, total ticket price (calculated as "number of people *
  unit price"), start time/end time of the visit;
- Dining: Restaurant name, total consumption cost, start time/end time of
  dining;
- Hotel: Name, cost (if applicable).
All content must be derived from real data returned by tools, and no
subjective generation is allowed.

## Cost and Rule Details
- Ticket Cost: Calculated based on the actual number of travelers;
- Hotel Cost: Calculated based on the number of rooms, with a maximum of 2
  people per room;
- Taxi Cost: Calculated according to the price displayed on the map for each
  question, regardless of the number of travelers;
- Bus Cost: 3 EUR per person;
- Walking/Driving: Cost is 0.

## Output Format Requirements (Final Result)
Only JSON format is allowed for output, and the outer layer must be wrapped
with the following json JSON must by included by json

### JSON Structure Example
json
{
  "itinerary": [
    {
      "date": "7.1",
      "schedule": [
        {
          "item": "hotel",
          "time": "8:00-8:00",
          "departure": "Name",
          "destination": "Name",
          "cost": 120,
          "transportation": "none",
          "referenceImage": "info.png"
        }
      ]
    }
  ]
}

You must output a single JSON object that strictly follows the itinerary
schema demonstrated in the example. Do not include any natural language
explanation, comments, or markdown outside the JSON. The top-level object
must contain an "itinerary" field, which is a list of day plans. Each day
plan must include a "date" and a "schedule" list.

json
{
  "itinerary": [
    {
      "date": "6.1",
      "schedule": [
        {
          "item": "hotel",
          "time": "8:00-8:00",
          "departure": "Pullman Paris",
          "destination": "Pullman Paris",
          "cost": 300,
          "transportation": "none",
          "referenceImage": "hotel_list.png"
        },
        {
          "item": "transportation",
          "time": "8:30-8:45",
          "departure": "Pullman Paris",
          "destination": "MUSÉE DU LOUVRE",
          "cost": 10,
          "transportation": "taxi",
          "referenceImage": "route_Pullman Paris_MUSÉE DU LOUVRE.png"
        },
        {
          "item": "attraction",
          "time": "09:30-13:30",
          "departure": "MUSÉE DU LOUVRE",
          "destination": "MUSÉE DU LOUVRE",
          "cost": 18,
          "transportation": "none",
          "referenceImage": "MUSÉE DU LOUVRE_guide.png,MUSÉE DU LOUVRE_ticket.png"
        },
        {
          "item": "transportation",
          "time": "13:30-13:36",
          "departure": "MUSÉE DU LOUVRE",
          "destination": "Les Antiquaires",
          "cost": 9.1,
          "transportation": "taxi",
          "referenceImage": "route_MUSÉE DU LOUVRE_Les Antiquaires.png"
        },
        {
          "item": "restaurant",
          "time": "13:45-15:15",
          "departure": "Les Antiquaires",
          "destination": "Les Antiquaires",
          "cost": 238,
          "transportation": "none",
          "referenceImage": "Les Antiquaires_guide.png,Les Antiquaires_reservation.png"
        },
        {
          "item": "transportation",
          "time": "15:15-15:26",
          "departure": "Les Antiquaires",
          "destination": "Musée d'Orsay",
          "cost": 10.1,
          "transportation": "taxi",
          "referenceImage": "route_Les Antiquaires_Musée d'Orsay.png"
        },
        {
          "item": "attraction",
          "time": "17:00-18:30",
          "departure": "Musée d'Orsay",
          "destination": "Musée d'Orsay",
          "cost": 188,
          "transportation": "none",
          "referenceImage": "Musée d'Orsay_guide.png,Musée d'Orsay_ticket.png"
        },
        {
          "item": "transportation",
          "time": "18:30-18:41",
          "departure": "Musée d'Orsay",
          "destination": "Le Meurice",
          "cost": 9.0,
          "transportation": "taxi",
          "referenceImage": "route_Musée d'Orsay_Le Meurice.png"
        },
        {
          "item": "restaurant",
          "time": "19:00-21:00",
          "departure": "Le Meurice",
          "destination": "Le Meurice",
          "cost": 350,
          "transportation": "none",
          "referenceImage": "Le Meurice_guide.png,Le Meurice_reservation.png"
        },
        {
          "item": "transportation",
          "time": "21:00-21:15",
          "departure": "Le Meurice",
          "destination": "Pullman Paris",
          "cost": 16.1,
          "transportation": "taxi",
          "referenceImage": "route_rutz_intercontinental.png"
        },
        {
          "item": "hotel",
          "time": "21:15-21:15",
          "departure": "Pullman Paris",
          "destination": "Pullman Paris",
          "cost": 0,
          "transportation": "none",
          "referenceImage": "hotel_list.png"
        }
      ]
    }
  ]
}

### Specific Format Requirements
1. The first and last items of the daily itinerary must be "hotel" (hotel)
   schedule, and the time must be filled in as "8:00-8:00";
2. A "transportation" (transportation) schedule must be inserted between
   every two different locations;
3. The date format is "month.day", e.g., "7.1" (August 1st);
4. The "item" field can only take the following four enumerated values:
   - "transportation"
   - "attraction"
   - "restaurant"
   - "hotel"
5. In the "transportation" segment, the departure location, destination,
   transportation method, and cost must be filled in; for other types
   (attraction, restaurant, hotel), the "transportation" field shall be
   filled in as "none";
6. The "referenceImage" (reference image) field must include the guide
   images and reservation images of the corresponding attraction or
   restaurant;
7. Transportation methods are limited to four types: "foot" (walking),
   "driving" (self-driving), "bus" (bus), and "taxi" (taxi);
8. For multi-day itineraries, the corresponding ticket booking interface and
   restaurant reservation interface must be called separately for different
   dates.
9. If the item is "attraction", "restaurant", or "hotel", then the departure
   and destination are both written as the same place.
\end{verbatim}
\end{Prompt}

\section{The WorldTravel-Webscape}
\label{sec:appendix_web_specs}

WorldTravel is not built on pre-structured databases or simplified tables. 
Instead, it introduces a large-scale, web-based environment that simulates how travel planning is performed in the real world: through diverse, visually rich webpages with mixed structured and unstructured content.

The environment contains over 2,000 \textbf{handcrafted webpages} spanning attractions, restaurants, hotels, transportation, booking systems, menus, and travel guides. 
Each task requires agents to navigate across multiple page types, extract decision-critical information from realistic interfaces, and integrate them into a globally feasible itinerary.

Unlike prior benchmarks that expose structured attributes directly, WorldTravel places all factual constraints, prices, availability time slots, and experiential cues inside rendered webpages. 
Agents must perceive complex webpages, extract decision-critical details, and integrate them into globally consistent itineraries.

\subsection{Design Principles of the WorldTravel-Webscape}

The WorldTravel-Webscape are constructed under three core principles:

\begin{itemize}[noitemsep, topsep=0pt]
    \item \textbf{Perception-grounded design.} 
    All task-relevant information is embedded in rendered webpages rather than exposed as structured fields. 
    Agents must visually parse layouts, recognize UI states (e.g., sold-out slots, disabled options, highlighted selections), and locate dispersed information across cards, tables, and panels.

    \item \textbf{Cross-page dependency.} 
    No single webpage is sufficient to solve a task. 
    Feasible planning requires jointly reasoning over multiple sources, such as attraction discovery pages, booking systems, transportation matrices, and travel guides, closely reflecting real-world travel planning workflows.

    \item \textbf{Template diversity.}
    Pages use varied layouts, interaction styles, and ways of presenting information, even within the same category. This prevents reliance on fixed templates and encourages agents to handle real, messy webpages.
\end{itemize}

\subsection{Attraction Overview Pages}

\begin{figure}[t]
    \centering
    \includegraphics[width=0.9\textwidth]{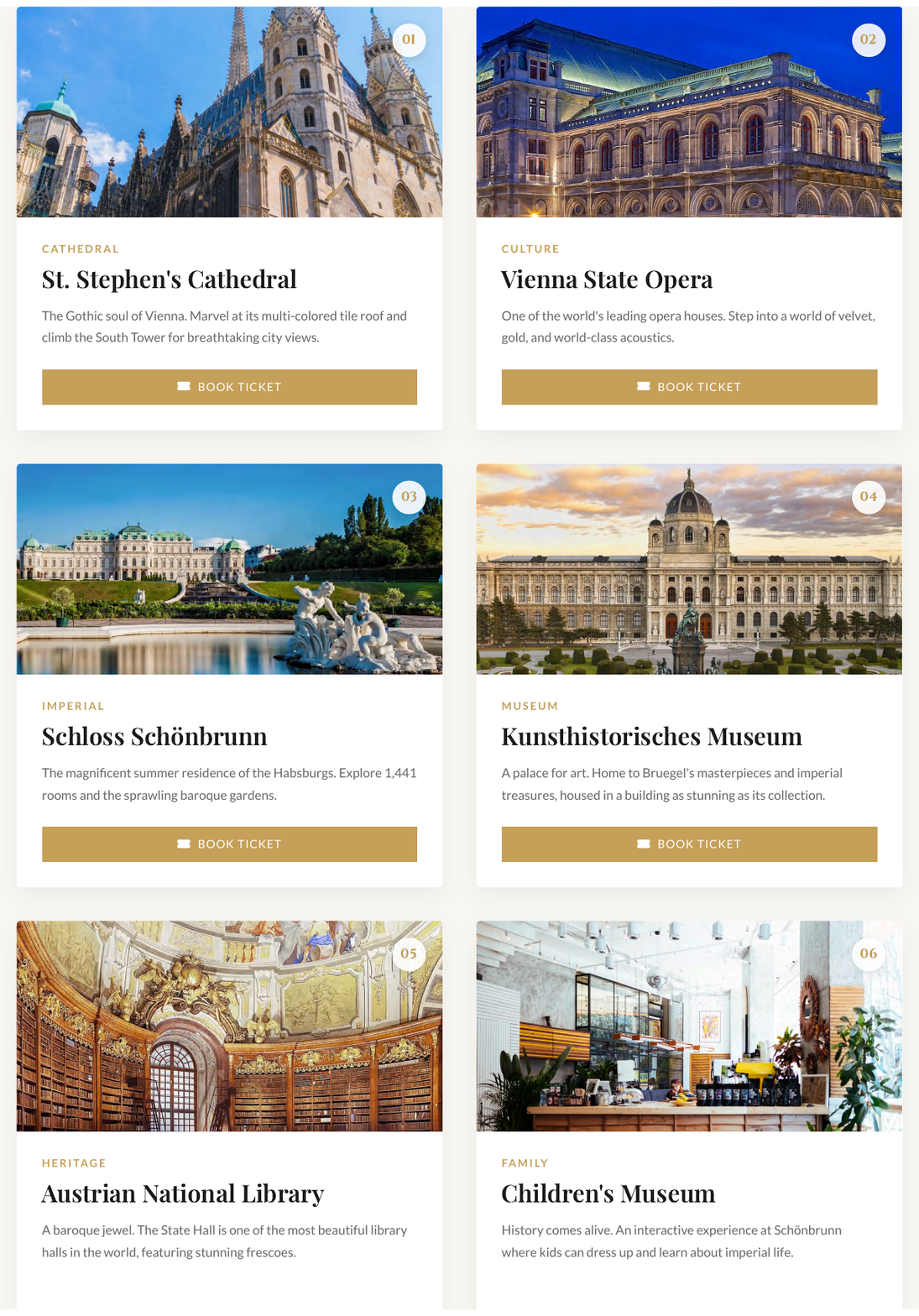}
    \caption{Attraction overview page displaying landmarks through category-tagged cards with imagery and descriptions.}
    \label{fig:attraction_overview}
\end{figure}

Attraction overview pages form the first layer of the environment (Figure~\ref{fig:attraction_overview}). They present landmarks, museums, and cultural sites through visually rich cards combining imagery, category tags (e.g., \emph{Culture}, \emph{Imperial}, \emph{Family}), and brief descriptions. 
These pages help narrow down what places are worth considering, such as art museums, family-friendly attractions, or scenic viewpoints, but they do not provide the details needed to schedule a visit. Information such as opening hours, ticket types, and reservation availability appears only on linked pages, so agents must move between different page types rather than rely on a single source.

\subsection{Restaurant Overview Pages}

\begin{figure}[t]
    \centering
    \includegraphics[width=0.9\textwidth]{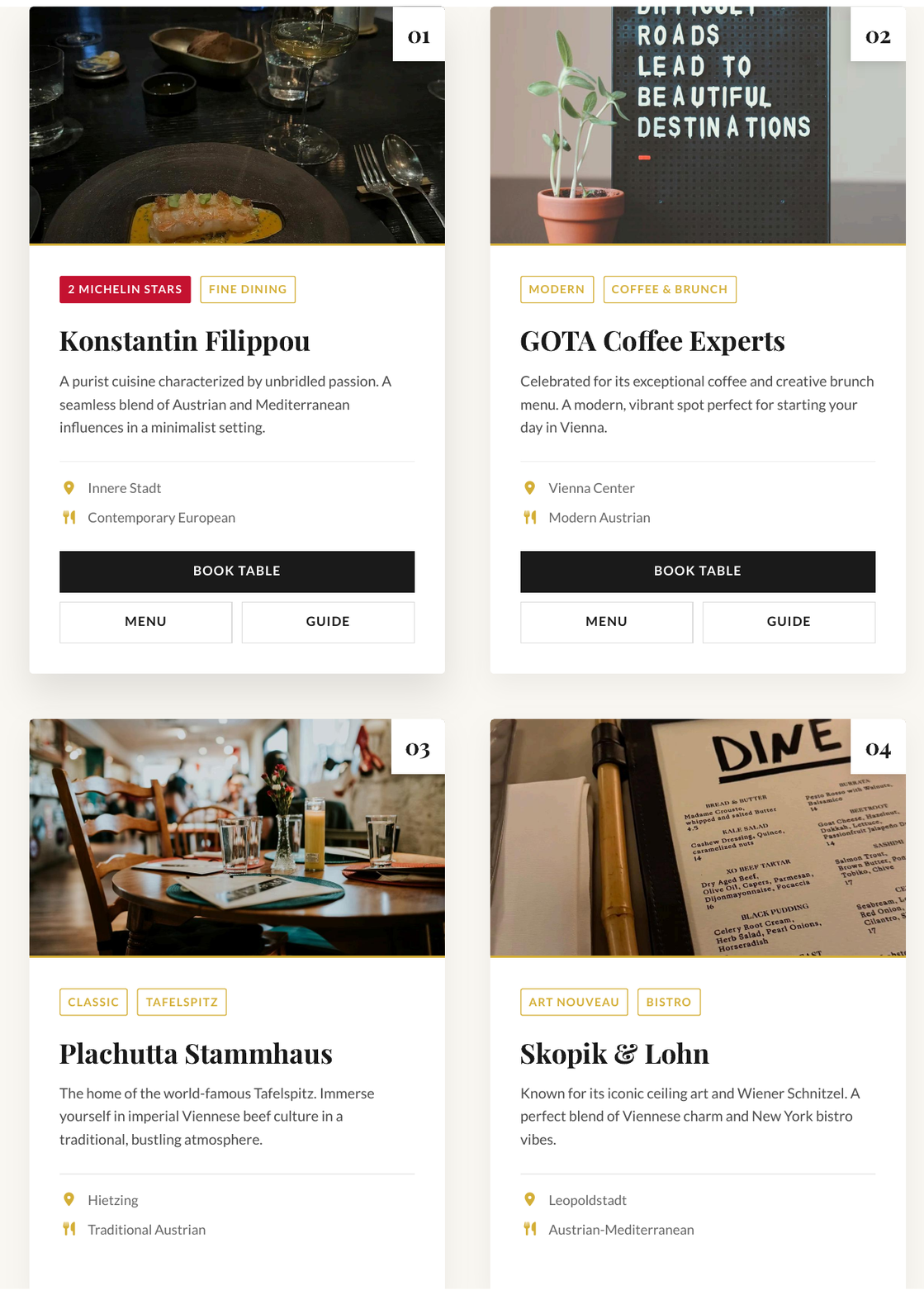}
    \caption{Restaurant overview page showing dining options with cuisine types, location context, and style indicators.}
    \label{fig:restaurant_overview}
\end{figure}

Restaurant overview pages bring together dining options of different styles and price ranges, from Michelin-star restaurants to casual cafés and bistros (Figure~\ref{fig:restaurant_overview}).
Each card shows images, cuisine type, location context, and links to reservations, menus, and dining guides.
These pages help users get a sense of what each place is like, but they do not show all the details needed to make a booking.
To decide whether a restaurant fits their needs (e.g., brunch, vegetarian, fine dining), agents must follow links to reservation pages or menus to check times, availability, and prices.

\subsection{Hotel Listing Pages}

\begin{figure}[t]
    \centering
    \includegraphics[width=0.9\textwidth]{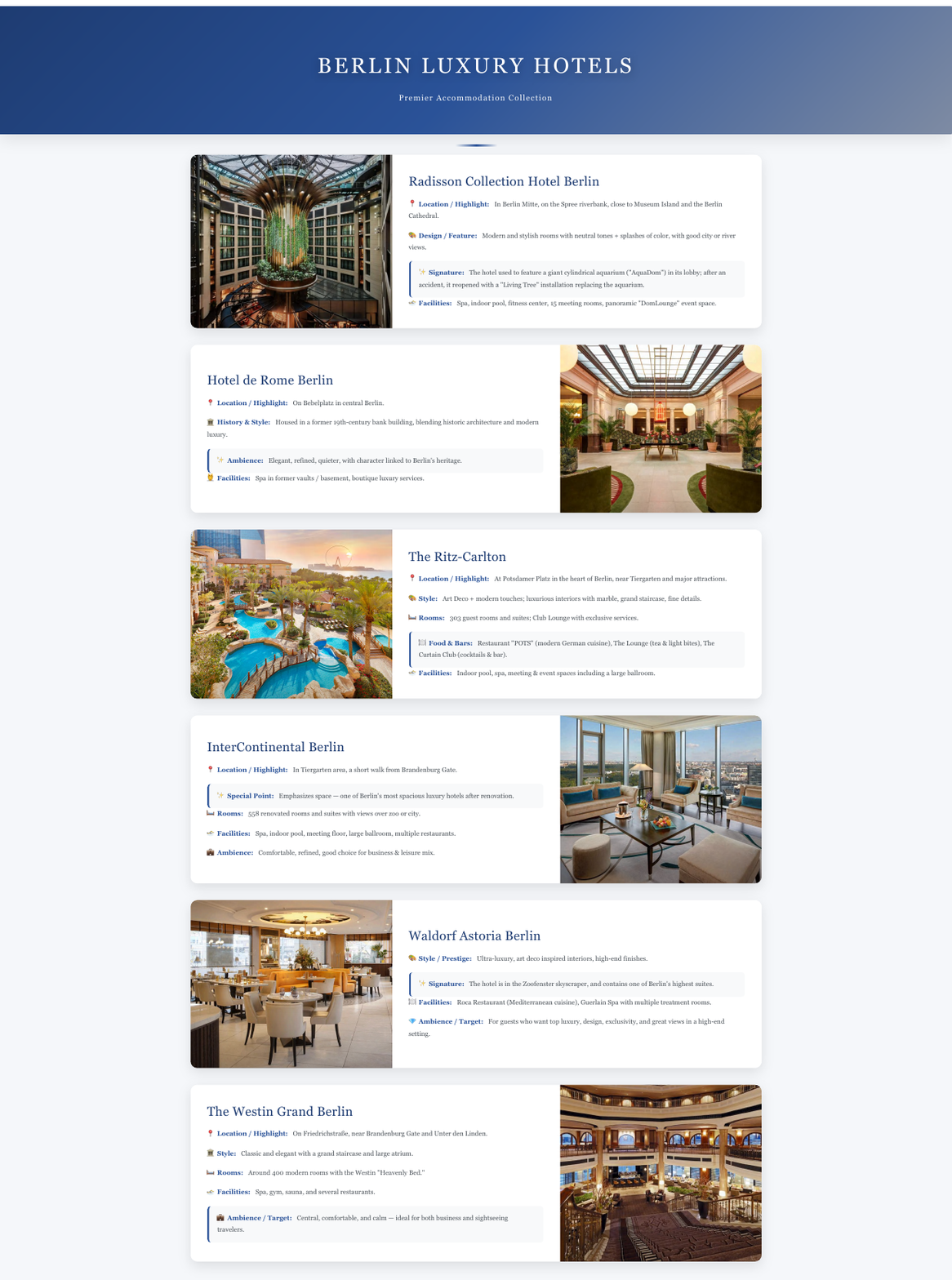}
    \caption{Hotel listing page presenting accommodations with editorial descriptions, amenities, and location highlights.}
    \label{fig:hotel_listing}
\end{figure}

Hotel listing pages present accommodation options in a side-by-side format, combining images, descriptive text, location notes, and amenity highlights (Figure~\ref{fig:hotel_listing}). Hotels are described through short editorial sections covering style, ambience, and background, instead of fixed attribute tables. To choose suitable hotels, agents must rely on these descriptions to match user preferences, such as comfort level, atmosphere, and location, rather than selecting from preset categories.

\subsection{Transportation Matrix Pages}

\begin{figure}[t]
    \centering
    \includegraphics[width=0.9\textwidth]{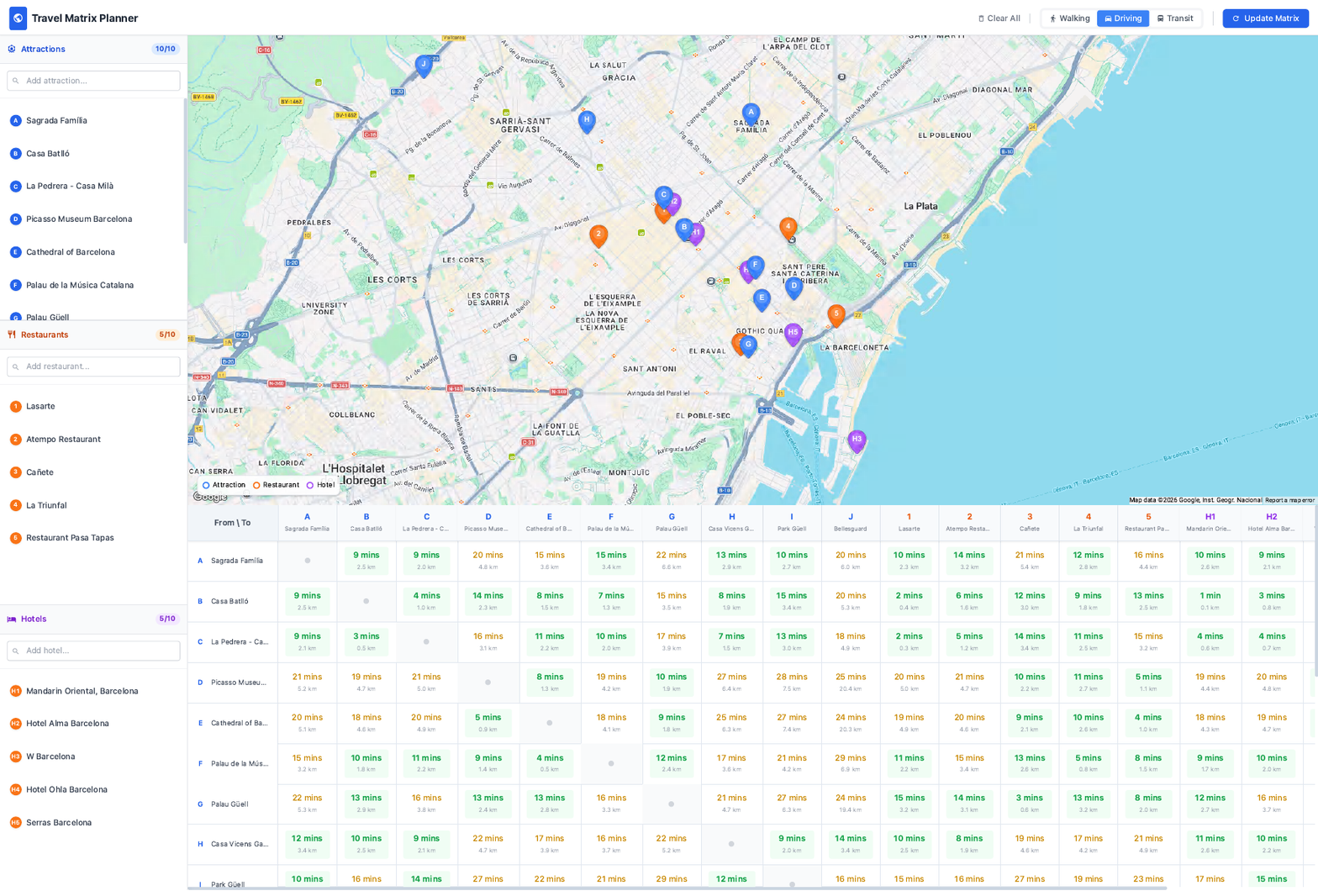}
    \caption{Transportation matrix page displaying pairwise travel times and distances across multiple modes.}
    \label{fig:transportation_matrix}
\end{figure}

Transportation matrix pages present travel times and distances between all attractions, restaurants, and hotels across multiple transportation modes (Figure~\ref{fig:transportation_matrix}).
They supply the core information needed to estimate transfers between locations.

\subsection{Attraction Booking Pages}

\begin{figure}[t]
    \centering
    \includegraphics[width=0.9\textwidth]{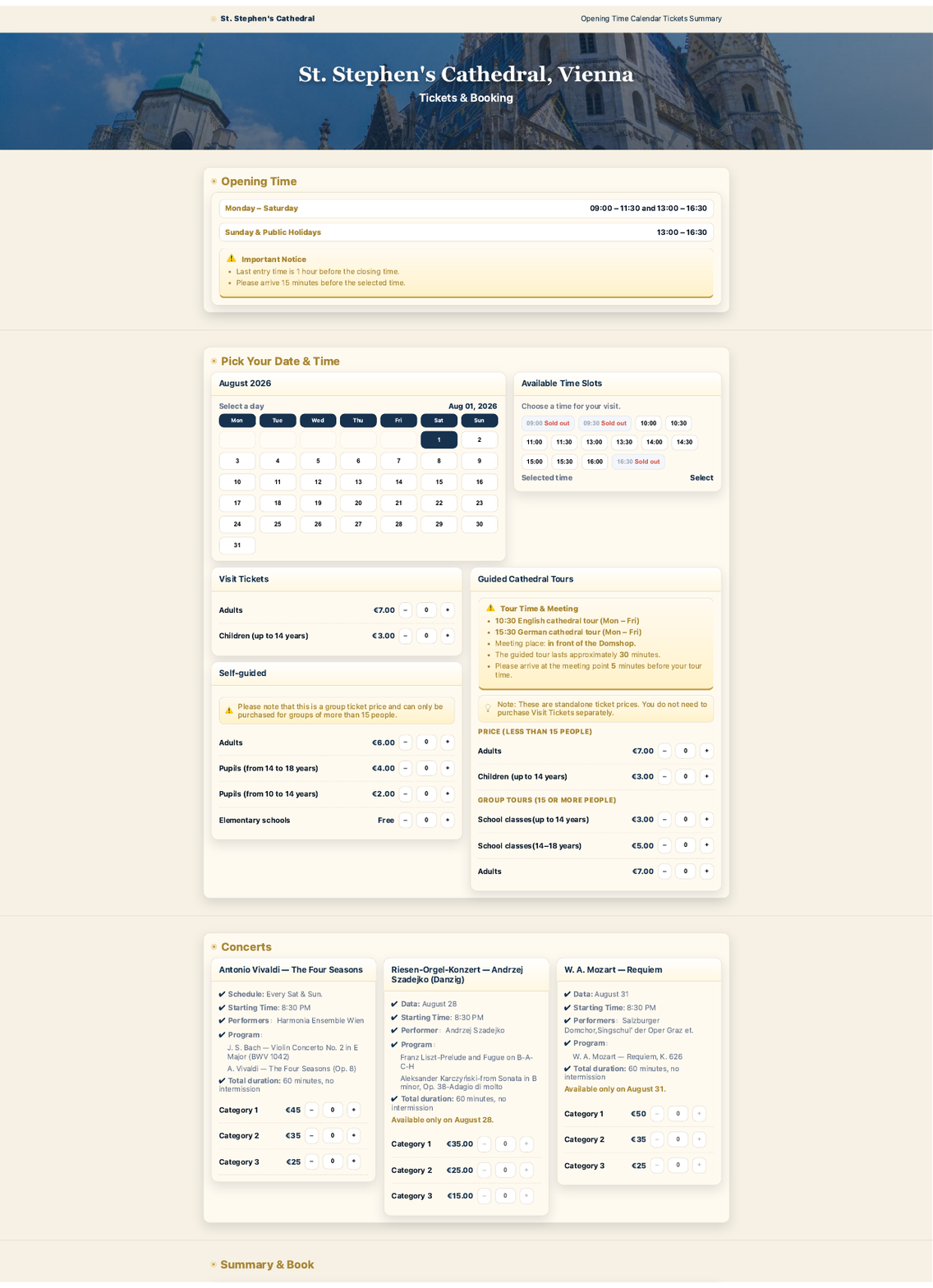}
    \caption{Attraction booking page with calendar, tiered pricing, visitor categories, and time slot availability.}
    \label{fig:attraction_booking}
\end{figure}

Attraction booking pages resemble real ticketing portals, featuring calendars, tiered ticket options, visitor categories, guided tours, and sold-out indicators (Figure~\ref{fig:attraction_booking}).
Fixed entry times, pricing rules, and arrival requirements are expressed through the interface rather than plain text.

\subsection{Restaurant Reservation Pages}

\begin{figure}[t]
    \centering
    \includegraphics[width=0.9\textwidth,height=0.95\textheight,keepaspectratio]{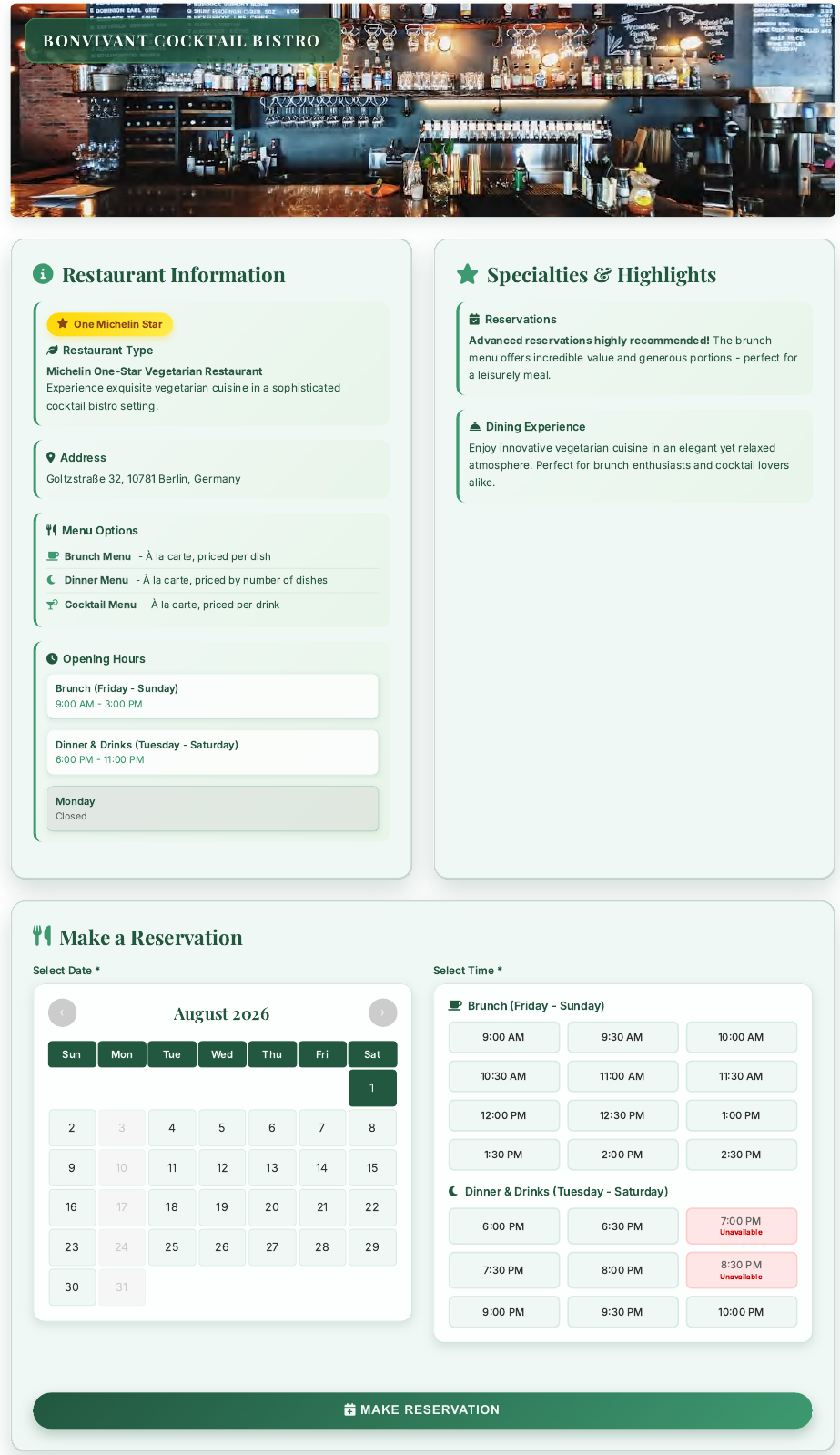}
    \caption{Restaurant reservation page with meal session calendars and time slot availability indicators.}
    \label{fig:restaurant_reservation}
\end{figure}

Restaurant reservation pages display calendars organized by meal sessions, with visual cues marking which time slots are available or sold out (Figure~\ref{fig:restaurant_reservation}).

\subsection{Attraction Travel Guides}

\begin{figure}[t]
    \centering
    \includegraphics[width=0.9\textwidth]{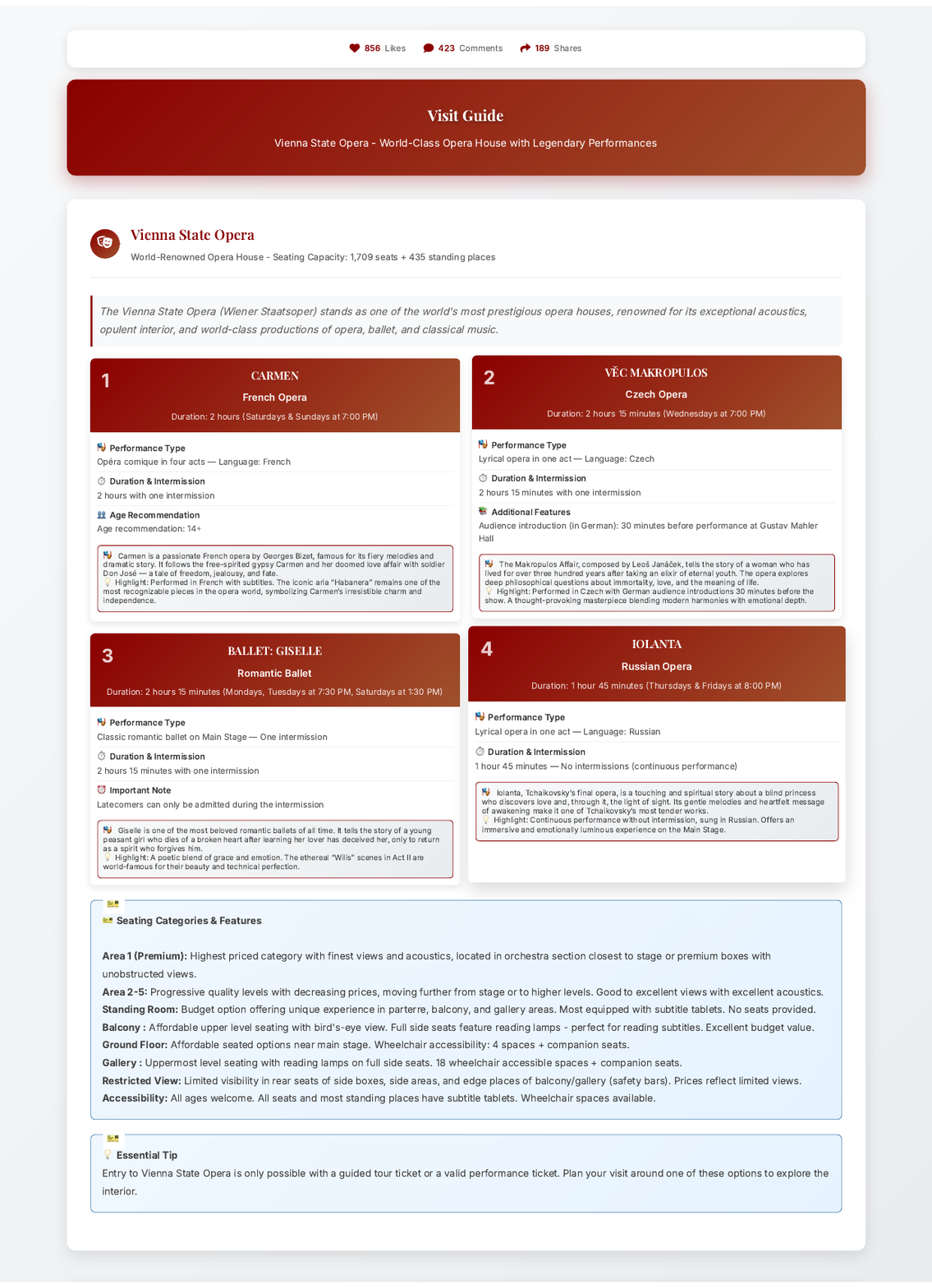}
    \caption{Attraction travel guide presenting visit recommendations in a blog-style format.}
    \label{fig:travel_guide}
\end{figure}

\begin{figure}[t]
    \centering
    \includegraphics[width=0.9\textwidth]{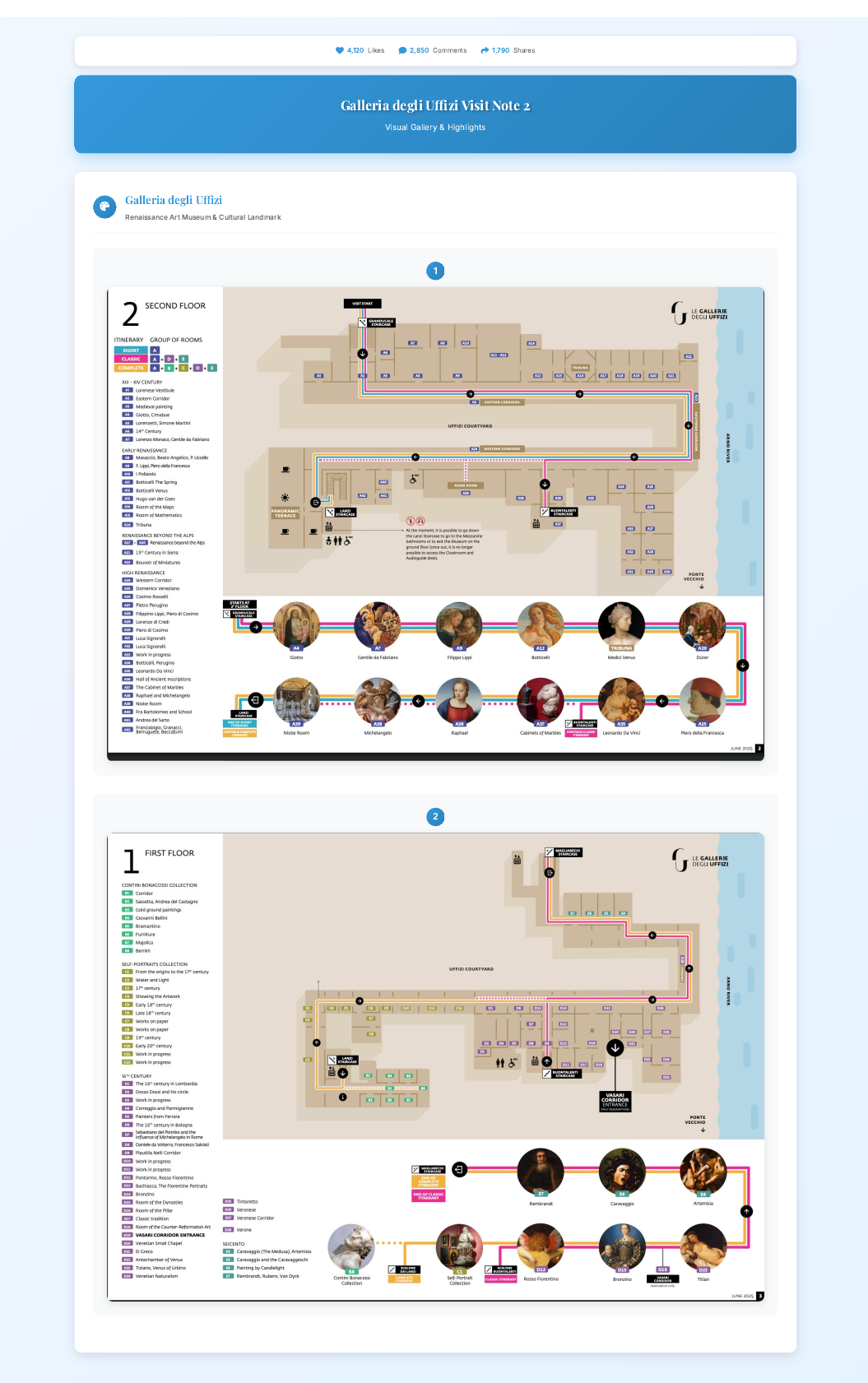}
    \caption{Scenic spot floor plan showing spatial layout and internal structure of an attraction.}
    \label{fig:scenic_spot_floor_plan}
\end{figure}

Attraction travel guides are written in the style of travel blogs or cultural features (Figure~\ref{fig:travel_guide}). Some attractions also include floor plan pages that display spatial layout and internal structure to help visitors navigate the venue (Figure~\ref{fig:scenic_spot_floor_plan}).

\subsection{Restaurant Menus}

\begin{figure}[t]
    \centering
    \includegraphics[width=0.9\textwidth]{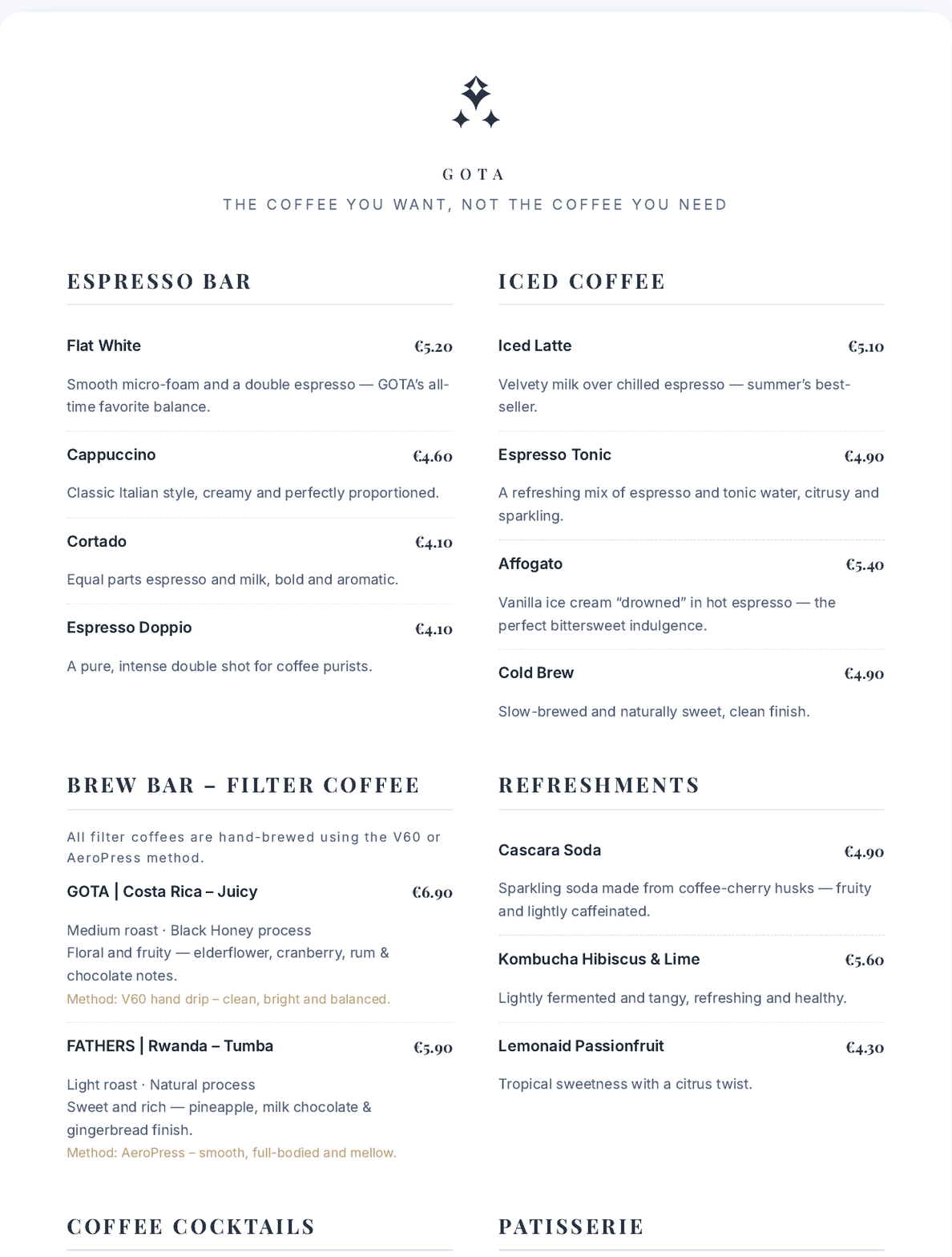}
    \caption{Restaurant menu page with dish listings, prices, and time-dependent offerings.}
    \label{fig:restaurant_menu}
\end{figure}

Restaurant menus present visually formatted listings of dishes, beverages, set meals, and time-dependent offerings such as weekend brunch (Figure~\ref{fig:restaurant_menu}).

\subsection{Restaurant Travel Notes}

\begin{figure}[t]
    \centering
    \includegraphics[width=0.9\textwidth]{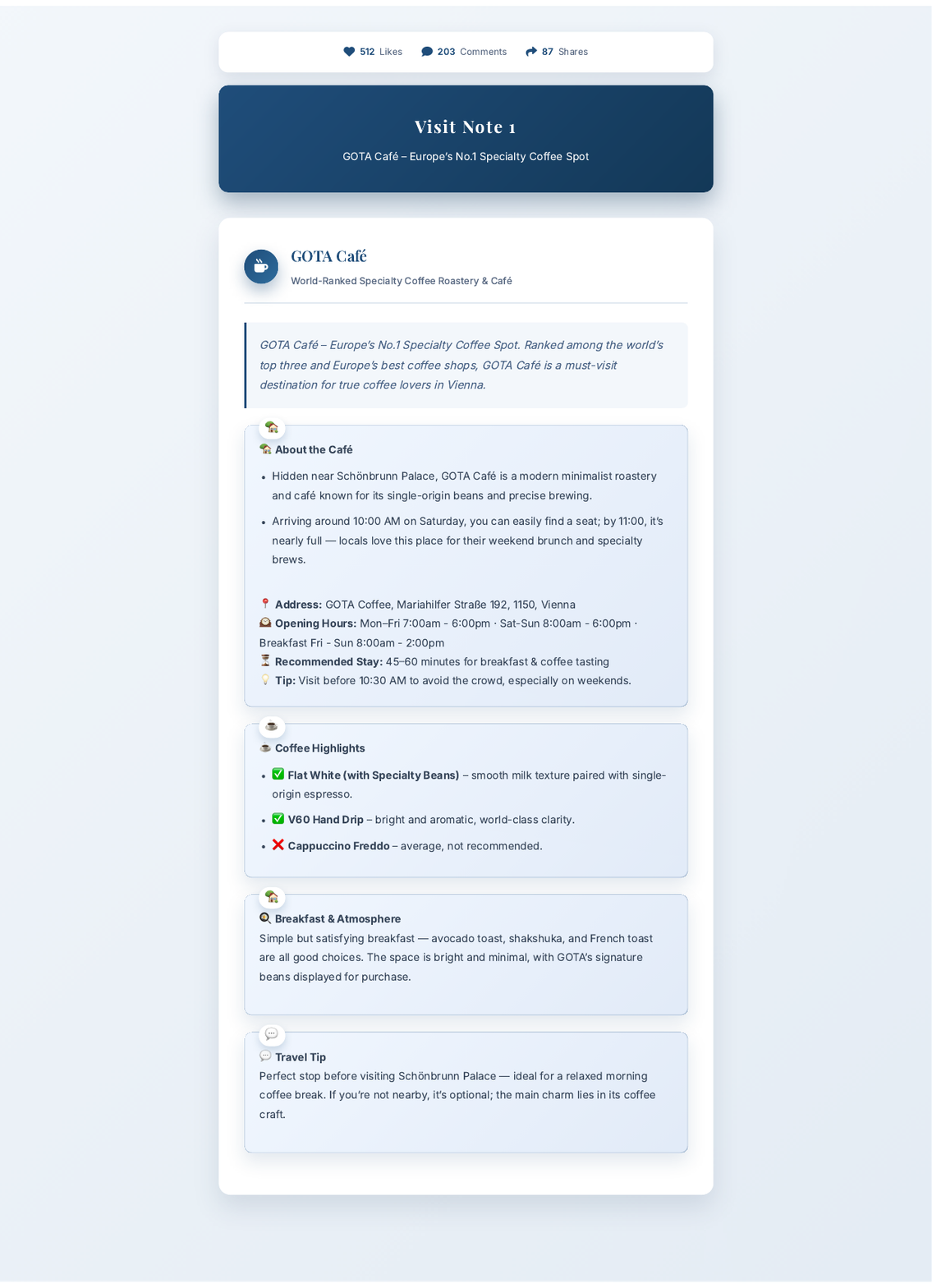}
    \caption{Restaurant travel notes combining subjective commentary with practical dining information.}
    \label{fig:restaurant_guide}
\end{figure}

Restaurant travel notes resemble user-generated travel blogs, combining subjective commentary with scattered factual details such as peak hours, recommended stay length, and best visiting times (Figure~\ref{fig:restaurant_guide}).

\end{document}